\documentclass[twoside,twocolumn,9pt]{article}
\usepackage{extsizes}
\usepackage[super,sort&compress,comma]{natbib}
\usepackage[version=3]{mhchem}
\usepackage[left=1.5cm, right=1.5cm, top=1.785cm, bottom=2.0cm]{geometry}
\usepackage{balance}
\usepackage{mathptmx}
\usepackage{sectsty}
\usepackage{graphicx}
\usepackage{lastpage}
\usepackage[format=plain,justification=justified,singlelinecheck=false,font={stretch=1.125,small,sf},labelfont=bf,labelsep=space]{caption}
\DeclareCaptionLabelFormat{tightlabel}{#1\,#2}
\usepackage{float}
\usepackage{fancyhdr}
\usepackage{fnpos}
\usepackage[english]{babel}
\addto{\captionsenglish}{%
  
}
\usepackage{array}
\usepackage{droidsans}
\usepackage{charter}
\usepackage[T1]{fontenc}
\usepackage[usenames,dvipsnames]{xcolor}
\usepackage{setspace}
\usepackage[compact]{titlesec}
\usepackage{hyperref}

\usepackage{epstopdf}
\graphicspath{{figures/}{head_foot/}}

\usepackage{draftwatermark}
\SetWatermarkText{DRAFT}
\SetWatermarkScale{1}
\SetWatermarkColor[gray]{0.8}

\definecolor{cream}{RGB}{222,217,201}

\begin{document}

\pagestyle{fancy}
\thispagestyle{plain}
\fancypagestyle{plain}{
\renewcommand{\headrulewidth}{0pt}
}

\makeFNbottom
\makeatletter
\renewcommand\LARGE{\@setfontsize\LARGE{15pt}{17}}
\renewcommand\Large{\@setfontsize\Large{12pt}{14}}
\renewcommand\large{\@setfontsize\large{10pt}{12}}
\renewcommand\footnotesize{\@setfontsize\footnotesize{7pt}{10}}
\makeatother

\renewcommand{\thefootnote}{\fnsymbol{footnote}}
\renewcommand\footnoterule{\vspace*{1pt}%
\color{cream}\hrule width 3.5in height 0.4pt \color{black}\vspace*{5pt}}
\setcounter{secnumdepth}{5}

\makeatletter
\renewcommand\@biblabel[1]{#1}
\renewcommand\@makefntext[1]%
{\noindent\makebox[0pt][r]{\@thefnmark\,}#1}
\makeatother
\renewcommand{\figurename}{Fig.}
\sectionfont{\sffamily\Large}
\subsectionfont{\normalsize}
\subsubsectionfont{\bf}
\setstretch{1.125} 
\setlength{\skip\footins}{0.8cm}
\setlength{\footnotesep}{0.25cm}
\setlength{\jot}{10pt}
\titlespacing*{\section}{0pt}{4pt}{4pt}
\titlespacing*{\subsection}{0pt}{15pt}{1pt}

\fancyfoot{}
\fancyfoot[LO,RE]{\vspace{-7.1pt}\includegraphics[height=9pt]{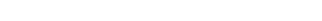}}
\fancyfoot[CO]{\vspace{-7.1pt}\hspace{13.2cm}\includegraphics{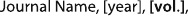}}
\fancyfoot[CE]{\vspace{-7.2pt}\hspace{-14.2cm}\includegraphics{head_foot/RF}}
\fancyfoot[RO]{\footnotesize{\sffamily{1--\pageref{LastPage} ~\textbar  \hspace{2pt}\thepage}}}
\fancyfoot[LE]{\footnotesize{\sffamily{\thepage~\textbar\hspace{3.45cm} 1--\pageref{LastPage}}}}
\fancyhead{}
\renewcommand{\headrulewidth}{0pt}
\renewcommand{\footrulewidth}{0pt}
\setlength{\arrayrulewidth}{1pt}
\setlength{\columnsep}{6.5mm}
\setlength\bibsep{1pt}

\makeatletter
\newlength{\figrulesep}
\setlength{\figrulesep}{0.5\textfloatsep}

\newcommand{\topfigrule}{\vspace*{-1pt}%
\noindent{\color{cream}\rule[-\figrulesep]{\columnwidth}{1.5pt}} }

\newcommand{\botfigrule}{\vspace*{-2pt}%
\noindent{\color{cream}\rule[\figrulesep]{\columnwidth}{1.5pt}} }

\newcommand{\dblfigrule}{\vspace*{-1pt}%
\noindent{\color{cream}\rule[-\figrulesep]{\textwidth}{1.5pt}} }

\makeatother

\twocolumn[
  \begin{@twocolumnfalse}
{\includegraphics[height=30pt]{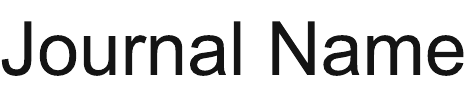}\hfill\raisebox{0pt}[0pt][0pt]{\includegraphics[height=55pt]{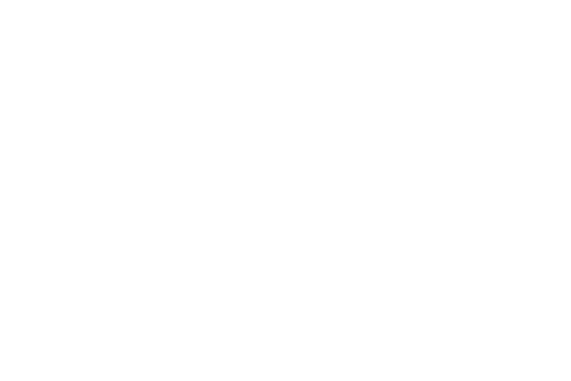}}\\[1ex]
\includegraphics[width=18.5cm]{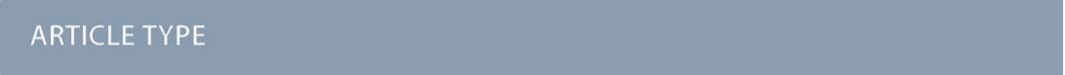}}\par
\vspace{1em}
\sffamily
\begin{tabular}{m{4.5cm} p{13.5cm} }

\includegraphics{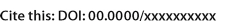} & \noindent\LARGE{\textbf{AEGIS: Assay-Aware Protocol Validation and Runtime Monitoring for Open-Source Liquid Handling Robots$^\dag$}} \\
\vspace{0.3cm} & \vspace{0.3cm} \\

 & \noindent\large{Priyanka V. Setty,$^{\ast}$\textit{$^{a,b}$} Arvind Ramanathan,\textit{$^{a}$} Ian Foster,\textit{$^{a,b}$} and Rick Stevens\textit{$^{a,b}$}} \\

\includegraphics{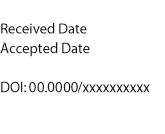} & \noindent\normalsize{Self-driving laboratories increasingly rely on low-cost liquid handling robots such as the Opentrons~OT-2, but these platforms ship without the pressure-based aspiration monitoring found on Hamilton or Tecan systems and are typically operated open-loop. Two complementary failure modes go undetected: protocols that are syntactically valid but violate assay-specific invariants (for example, tip reuse between PCR template and a no-template control), and physical execution failures (partial dispense, air bubbles, missing tips) that occur at runtime. We present AEGIS, a two-layer guardian that addresses both. Layer~1 combines a curated machine-readable assay rule database with a large language model that reasons over OT-2 Python code, reaching an adjusted F\textsubscript{1} of 0.97 on a 24-protocol benchmark spanning five assay families and outperforming both rules-only and LLM-only ablations across five backends; a free open-weight model ties the best proprietary one, so the approach does not depend on a paid API. Layer~2 fits a principal-component world model to YOLO-cropped four-frame pipette trajectories and detects physical anomalies on the p1000 channel; under a leakage-free leave-one-plate-out evaluation it reaches an average precision of 0.89 and an operating-point F\textsubscript{1} of 0.71 (AUROC 0.80), a deployment-faithful figure that matches the live demonstration, and we characterize the small-pipette (p20) resolution limit at which the front view degrades (F\textsubscript{1}=0.47). A proof-of-concept live demonstration on a physical OT-2 with a front-view camera (five replicates per condition) catches the planted no-tip failure deterministically (5/5) and partial dispense under the default cascade on red and yellow dye (3/5 each); an always-VLM operating point that uses a three-assessment self-vote prompt lifts partial-dispense recall to 5/5 on both red and yellow. Water remains a principled limit of any front-view-only system (cascade 0/5, always-VLM 1/5 at low confidence), which AEGIS surfaces in the VLM's own low-confidence reasoning rather than over-committing to a verdict. End-to-end the OT-2 halts at the next protocol-level boundary at an effective-halt latency of $13.27 \pm 0.47$\,s ($n{=}9$ paused runs), and the cascade triage holds projected per-plate VLM cost at \$1.63 versus \$10.33 for an always-VLM baseline. AEGIS is open source and is, to our knowledge, the first system to unify pre-flight assay-aware validation with runtime visual monitoring for an open-source liquid handler.
} \\

\end{tabular}

 \end{@twocolumnfalse} \vspace{0.6cm}

  ]

\renewcommand*\rmdefault{bch}\normalfont\upshape
\rmfamily
\section*{}
\vspace{-1cm}


\footnotetext{\textit{$^{a}$~Data Science and Learning Division, Argonne National Laboratory, Lemont, IL 60439, USA. E-mail: psetty@anl.gov}}
\footnotetext{\textit{$^{b}$~Department of Computer Science, University of Chicago, Chicago, IL 60637, USA.}}

\footnotetext{\dag~Electronic Supplementary Information (ESI) available: rule database, per-protocol breakdowns, fleet scaling logs, demo run videos. See DOI: 00.0000/00000000.}



\section{Introduction}
\label{sec:introduction}

\begin{figure}[!t]
  \centering
  \includegraphics[width=\columnwidth]{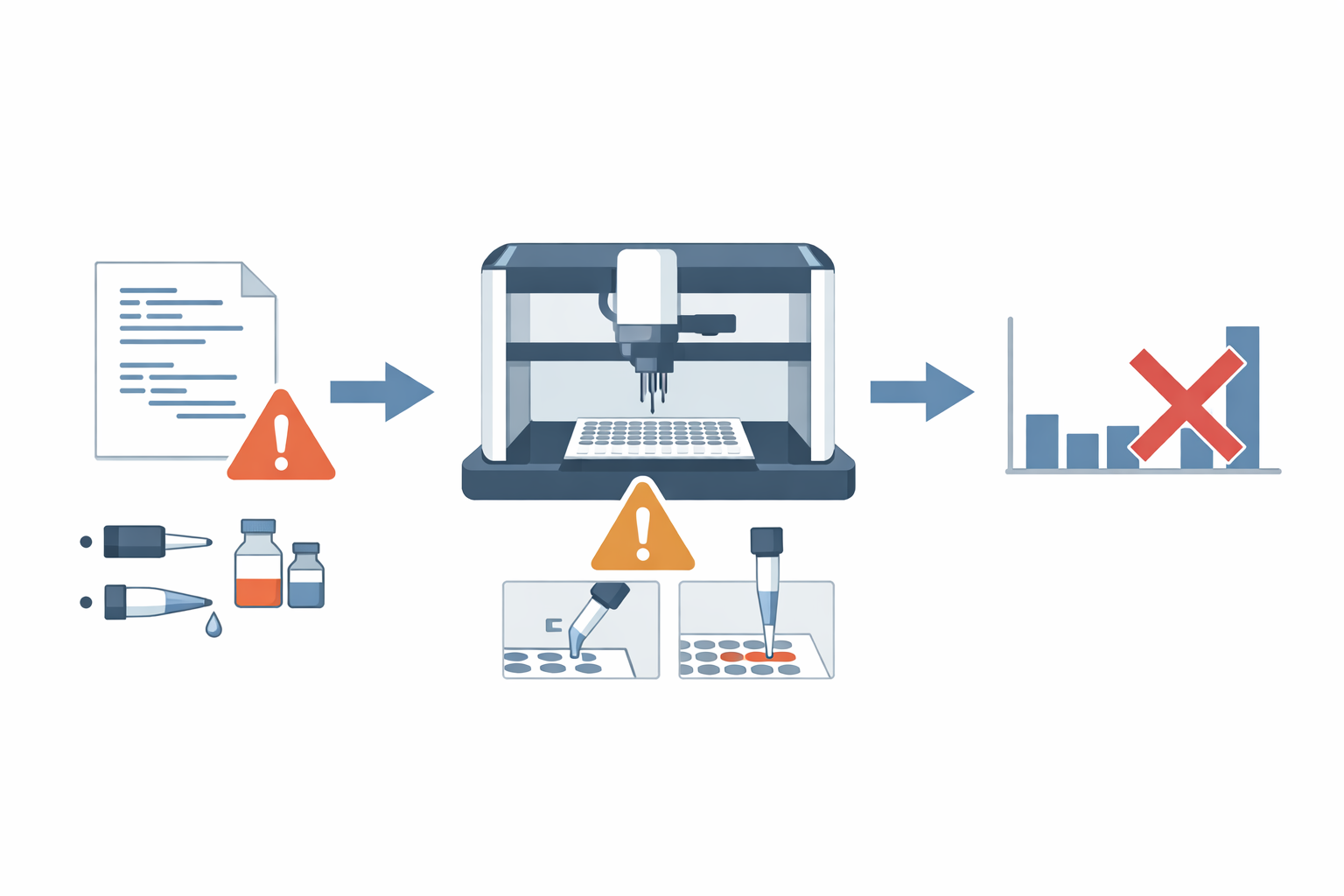}
  \caption{AEGIS guards against two silent failure modes that corrupt autonomous-lab data: assay-level (semantic) errors in the protocol code, and physical execution failures inside the liquid handler during the run. Both propagate as bad data into downstream analyses and the predictive models trained on them.}
  \label{fig:intro-problem}
\end{figure}

Self-driving laboratories (SDLs) are reshaping experimental science by closing the design--build--test--learn loop with autonomous execution: predictive models propose experiments, robotic platforms run them, and the resulting data updates the model in a continuous loop, ideally without human intervention.\cite{SDL2025Review,ALab2023,Coscientist2023} The scientific integrity of this loop depends on whether each wet-lab step actually does what the protocol intended; at scale, no human can audit every well to verify. A silently failed wet-lab run does not produce an error; it produces \emph{corrupted data that compounds} across subsequent model updates and experimental suggestions. Yet across the SDL stack, the robots themselves are \emph{biologically blind}: a liquid handler executes whatever Python or vendor-specific protocol it is given, with no awareness of assay-level intent. Two distinct failure modes can silently corrupt a run (Fig.~\ref{fig:intro-problem}): assay-level violations that survive every syntactic check, and physical execution failures that the robot's own log will not flag. The A-Lab\cite{ALab2023} synthesized 41 of 58 target compounds (a 71\% success rate), a vivid reminder that ``the robot ran'' is not the same as ``the experiment worked.''

\subsection{Assay-level (semantic) errors}
\label{sec:intro:semantic}

An assay-aware reading of the protocol's Python source catches errors that no syntactic check can: the tip touching a no-template control should never have touched a PCR template,\cite{Kwok1989} standards in a Bradford protein assay\cite{Bradford1976} must be dispensed low-to-high to limit carryover, and in a high-throughput drug-screening assay\cite{Zhang1999} a control well must never receive the drug under test. A protocol can be syntactically valid, accepted by the vendor's simulator, and executed without runtime errors, yet still corrupt every downstream measurement through such an assay-level (semantic) violation.

A wave of LLM agents now converts natural-language experimental intent into executable protocols,\cite{Coscientist2023,LabscriptAI2025,Opentrons2024} but none of them enforces the per-assay invariants a wet-lab scientist would apply on a code read. Coscientist couples GPT-4 with web search and OT-2 control, correcting its own coding errors by consulting the equipment manual at runtime; the error handling is reactive code correction, not pre-flight assay validation.\cite{Coscientist2023} LabscriptAI generates liquid-handling scripts from natural language across multiple platforms including the OT-2, and was used in production for 298 GFP variants across 53 teams, but its 55-task benchmark measures protocol-generation success and executability, not assay-specific rules.\cite{LabscriptAI2025} Opentrons' commercial AI tool pairs a retrieval-augmented generator with the Protocol Designer to produce executable code, also without assay-level validation.\cite{Opentrons2024} Adjacent work (ChemCrow's tool-augmented chemistry agent with built-in safety screening,\cite{ChemCrow2024} ORGANA's GPT-4-driven robotic chemistry assistant that turns natural-language goals into syntactically verified robot plans with human-in-the-loop troubleshooting,\cite{ORGANA2024} CLAIRify's verifier-assisted translation of natural-language protocols into an executable robot description language,\cite{CLAIRify2024} and BioPlanner's benchmark of LLM protocol planning in biology\cite{BioPlanner2024}) each advances protocol authoring or safety screening, but none encodes the per-assay biology invariants (fresh tips between PCR template and NTC, low-to-high standards in Bradford, mix steps after each serial dilution) that distinguish a correct biology protocol from a merely-compiling one. The closest formal-methods precedent is BioScript, a domain-specific language that statically verifies against unsafe chemical mixing on microfluidic laboratories-on-a-chip,\cite{BioScript2018} but its scope is microfluidics, not bench-scale liquid handling. The unaddressed gap across this body of work is consistent: existing systems check that a protocol is syntactically and logically well-formed, but none reads it through the lens of assay-specific biology invariants.

PRISM, a pre-flight planning and validation system, checks the biological plausibility of a proposed protocol and simulates inter-robot plate transfers in an Omniverse-based digital twin before execution.\cite{PRISM2026} AEGIS differs along two axes. First, where PRISM judges biological plausibility at the level of the overall protocol, Layer~1 checks biological correctness against \emph{explicit, assay-specific rules}, the finer and auditable per-assay invariants described above (for example, a tip carried from a PCR template into a no-template control). Second, the digital twin captures the \emph{physical movements} of the workflow, gantry motion and plate transfers between instruments, but not the liquid handler's internal pipetting; aspiration and dispensing inside the tip lie outside the simulated scene, and the observational window ends at the start of the run, the runtime gap Layer~2 fills.

\subsection{Physical execution errors}
\label{sec:intro:physical}

Even a correct protocol can fail at runtime through purely physical events: a missing tip, an air bubble drawn during aspirate, accumulated residual liquid in a reused tip, or a missed dispense. These failures are invisible to the protocol code and require runtime observation of the robot itself.

At runtime, industrial liquid handlers from Hamilton and Tecan ship with pressure-based monitoring.\cite{Hamilton2024,Tecan2024} Hamilton's Total Aspiration and Dispense Monitoring (TADM) records a pressure-versus-time trace for every pipetting channel and compares it against expected per-liquid-class envelopes. Tecan combines capacitive (cLLD) and pressure-based (pLLD) liquid-level detection with Process Monitored Pipetting for similar coverage. These capabilities ship only on platforms in the \$100{,}000--\$300{,}000 class with proprietary scripting and orchestration software (Hamilton Venus, Tecan Evoware), an integration profile incompatible with the open, Python-native, multi-robot scenarios most academic SDLs target. The Opentrons~OT-2, the most widely deployed open-source liquid handler in academic SDLs,\cite{LabscriptAI2025,PRISM2026,Opentrons2024} ships with no aspiration or dispense monitoring, no liquid-level detection, and no error reporting beyond Python exceptions. Any runtime quality control for the OT-2 must therefore be added externally, typically through cameras or auxiliary sensors.

Vision-based monitoring for the OT-2 has so far focused on narrow detection tasks. A YOLOv8 pipeline has been deployed on the OT-2 to detect pipette tips and liquid volumes in real time, with the authors explicitly noting inability to detect air bubbles as future work.\cite{DTU2025} An attention-augmented YOLOv8 variant targets liquid retention (residual droplets) on tips across different liquid colours and backgrounds, but is restricted to per-frame classification of a single failure type rather than analyzing the trajectory across the pipetting cycle.\cite{LiquidRetention2024} The LIRA edge module uses vision-language models for localisation and post-action inspection in SDLs and reports a ${\sim}10\times$ reduction in localisation time, but is fundamentally reactive: it inspects \emph{after} each action without a model of what the protocol expected to happen, treating every step as an independent inspection task with no shared prior across the run.\cite{LIRA2025}

A separate body of work uses learned world models as the basis for anomaly detection. The prediction error of a model-based RL world model has been shown to serve as a domain-agnostic anomaly signal on locomotion and flight tasks.\cite{WorldModelAnomaly2025} Hybrid frameworks combine supervised failure classification with world-model anomaly thresholds for visual inspection of industrial gauges.\cite{WorldModelFailure2026} The dominant world-model architectures (DreamerV3,\cite{DreamerV3} IRIS,\cite{IRIS2023} TD-MPC2\cite{TDMPC2024}) are evaluated on games, locomotion, or tabletop manipulation. MATTERIX provides a GPU-accelerated digital twin of chemistry lab automation built on Isaac~Sim/Lab,\cite{MATTERIX2025} but its purpose is in-silico workflow testing rather than runtime anomaly detection. Methodologically the closest precedent is industrial visual anomaly detection (MVTec~AD,\cite{MVTec2019} PatchCore\cite{PatchCore2022}), which trains on normal images alone and flags deviations. None of these world-model or train-on-normal anomaly approaches has been applied to liquid handling, which is the gap AEGIS Layer~2 closes by adapting the train-on-normal paradigm to temporal pipette trajectories on the OT-2.

A parallel and faster-moving line scales foundation and vision-language-action (VLA) models to laboratory and embodied automation. RoboChemist couples vision-language models with a robotic agent to carry out multi-step chemistry protocols while monitoring compliance with experimental norms,\cite{RoboChemist2025} and recent surveys catalogue how foundation models are reshaping both high-level task planning and low-level control for robot manipulation, together with their scalability and safety challenges.\cite{EmbodiedFMSurvey2025} These systems learn rich, general policies from large datasets and substantial compute.
AEGIS deliberately takes the opposite, lightweight and data-efficient route appropriate to a proof of concept: a train-on-normal anomaly monitor and an explicit rule checker that need no large training corpus and run on commodity hardware. We see the foundation-model path as the natural scale-up, a fine-tuning target once monitoring datasets grow large enough to cover the failure modes AEGIS does not yet handle, and we return to it in the future-work discussion (Sec.~\ref{sec:conclusion}).

\subsection{Contributions and goals}
\label{sec:intro:contributions}

We address both gaps on the Opentrons~OT-2. The OT-2 exposes its protocols as Python and integrates easily with multi-robot SDL workflows through its open HTTP API, but ships with no built-in tip-level or pressure-based sensing. We present \textbf{AEGIS} (Automated Experiment Guardian and Inspection System), a proof-of-concept two-layer guardian that runs end-to-end on a physical OT-2 with off-the-shelf hardware: a front-view USB camera, a laptop, and the existing Opentrons API.

\textbf{Layer~1} is a pre-flight assay-aware checker that reads the protocol's Python source and flags \emph{violations of a curated 22-rule database} of assay-specific invariants spanning five families (contamination, wrong well order, missing or out-of-order steps). We frame its evaluation as a \emph{rule-coverage benchmark}, whether a violation of each formalized rule is detected, rather than a claim of detecting arbitrary or unseen semantic errors, and we assess both frontier and freely available open models as the underlying reasoner.

\textbf{Layer~2} is a runtime monitor that captures the pipette trajectory from the front-view camera, detects silent physical failures (partial dispense, no aspirate, no tip) as they occur, and pauses the OT-2 via its API to notify the operator; it adapts the train-on-normal anomaly-detection paradigm from industrial visual inspection, using a lightweight PCA reconstruction over temporal four-frame pipette trajectories. Because its signal is the tip's geometric trajectory rather than reagent appearance, detection does not depend on reagent colour; truly transparent fluids such as water are the exception, a physical limit of a front-view monitor that we report honestly (Section~\ref{sec:results:demo}). The PCA scorer is a deliberately simple, data-efficient baseline for the small-data regime of a proof of concept, designed to scale as deployments accumulate trajectories (Section~\ref{sec:conclusion}).

We demonstrate AEGIS end-to-end on a physical OT-2 across multiple failure modes, characterizing where the system catches reliably, where it currently falls short, and what extensions would close the remaining gaps. We release the validators, rule database, trained model weights, captured datasets, and demonstration protocols under an open-source license.

\section{Methods}
\label{sec:methods}

\begin{figure}[!t]
  \centering
  \includegraphics[width=\columnwidth]{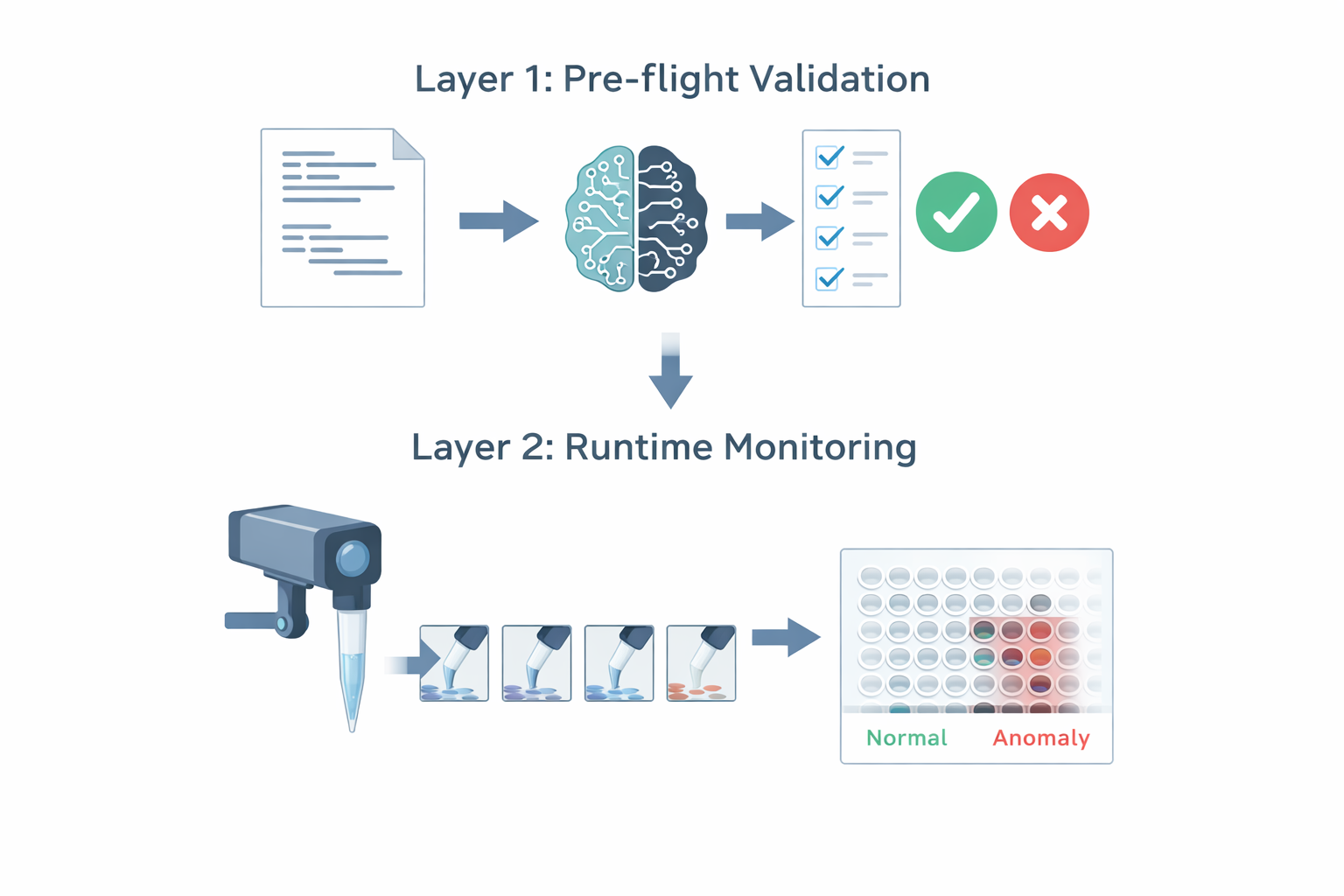}
  \caption{AEGIS two-layer architecture. \textbf{Layer~1} reads the OT-2 protocol's Python source before the run and flags assay-level violations against a curated rule database. \textbf{Layer~2} watches the pipette through a camera during execution, scores per-well trajectories with a PCA world model trained on normal trajectory images, and routes ambiguous wells to a vision-language model that arbitrates the pause decision.}
  \label{fig:architecture}
\end{figure}

AEGIS is a two-layer guardian for the Opentrons~OT-2 (Fig.~\ref{fig:architecture}): a pre-flight assay-aware validator (Layer~1) that reads the protocol's Python source before the run, and a runtime visual monitor (Layer~2) that watches the pipette through a camera, scores per-well trajectories with a PCA world model trained on normal trajectory images, and routes ambiguous wells to a vision-language model (VLM) that arbitrates pause decisions. Both layers run as a single per-robot agent and can be replicated across a fleet without code changes; the multi-robot scaling experiment is reported in the supplementary information (SI~Section~A). Source code, weights, and datasets are released under the MIT license (see Conclusions).

\subsection{Protocol benchmark and failure injection}
\label{sec:methods:benchmark}

We curated 24 OT-2 Python protocols spanning five assay families: serial dilution, polymerase chain reaction (PCR), Bradford protein assay, cell-treatment with drug controls, and generic transfer/normalization. Of these, 11 are correct (drawn from Opentrons' official tutorials and the Protocol Library, with minor formatting cleanup) and 13 are buggy variants in which an assay-relevant error has been injected: a shared tip between PCR template and the no-template control, a missing master-mix step, an inverted serial-dilution direction, a high-to-low Bradford standards series, and similar. Each buggy protocol tests exactly one rule, simplifying per-rule evaluation. A separate set of three multi-bug protocols (each with two or three concurrent errors) is used to test whether the validator surfaces multiple independent violations from a single code read. We feed every protocol through Layer~1 and compare the validator's predicted pass/fail and per-rule call against the injected ground truth (Section~\ref{sec:results:l1}). Two of the eleven nominally-correct tutorial protocols in fact contain a genuine tip-reuse bug; we therefore report both a strict score against the original labels and an adjusted score that relabels these two as buggy, crediting the models that flag them.

\subsection{Layer 1: assay rules and validator}
\label{sec:methods:l1}

We formalized 22 assay-specific rules in a machine-readable JSON schema. Each rule has an identifier (e.g.\ \texttt{PCR-2}: ``fresh tip between every PCR template and the NTC''), a natural-language statement, a severity level, an assay scope, and optional structural indicators the LLM uses during code inspection. Rules cover contamination (tip-reuse patterns), ordering (low-to-high standards, dilution direction), completeness (master-mix presence, mixing after each dilution step), and source/destination correctness (drug only in treated wells, diluent from the correct reservoir). The rule database is extensible by design: adding an assay family requires only editing the rules file, with no code changes to the validator.

Layer~1 itself is a single-pass LLM call that consumes the rule database, a curated Opentrons API reference describing the semantics of \texttt{transfer}, \texttt{distribute}, \texttt{consolidate}, and the \texttt{new\_tip} parameter, and the protocol's Python source. We supply this Opentrons API reference in-context rather than relying on each backend's pretrained knowledge of the library, so the comparison reflects assay-level reasoning rather than familiarity with Opentrons syntax and does not disadvantage the open-weight model. The LLM is asked in one structured prompt for (i)~the inferred assay family with a confidence label, (ii)~a rule-by-rule evaluation that lists offending wells and cites each violated rule, and (iii)~a final pass/fail verdict. We evaluate five backends spanning capability tiers: OpenAI o4-mini (reasoning), Claude Opus~4 and GPT-4o (frontier), Claude Sonnet~4.5 (mid-tier), and NVIDIA Nemotron-3 Ultra 550B (open-weight, served through NVIDIA NIM). The open-weight model is included deliberately to test whether assay-aware validation depends on a paid frontier API or can run on openly available weights, a deployment-cost question for self-driving laboratories. All backends are scored on the same 24 single-bug protocols; the multi-bug set is a separate test reported below. Combining classification and rule checking in a single call lets the model reason about cross-step context, for example that a loop's \texttt{transfer} call inherits the \texttt{new\_tip} default, which a staged pipeline would lose. The design is deliberately hybrid: rules tell the LLM \emph{what} to check; the LLM figures out \emph{how} to check it in the code. To isolate each component's contribution we evaluate Layer~1 in three configurations: the full hybrid (rules + LLM), an LLM-only ablation (the same prompt and Opentrons API reference, with only the rule database withheld, so the ablation measures the rules' contribution and not library knowledge), and a deterministic rules-only baseline (structural pattern-matching without the LLM). For the open-model comparison we also record per-protocol wall-clock latency for each backend.

\subsection{Layer 2 setup and data collection}
\label{sec:methods:l2:data}

A Logitech webcam mounted on a stand pointed at the OT-2 deck captures the work area at 1\,fps, with a plain black cardboard panel placed behind the deck to simplify tip segmentation against a uniform background. The live monitor reads the Opentrons HTTP API to obtain each pipetting command's completion timestamp and aligns the corresponding captured frame; alignment is verified to within $\pm 1$ frame. Cross-clock drift between the OT-2 and the workstation is auto-detected at run start from the HTTP response \texttt{Date} header (typically ${\sim}25\,$s in our setup) and applied to every command timestamp before frame mapping.

For the offline evaluation we collected 64 p1000 well trajectories on the same hardware: 26 physically-normal wells (1--2 from the early wells of each plate) and 38 planted failures, across three reagent contrasts (red, yellow, water) and four failure types (partial dispense, no dispense, no tip, air bubble). Each trajectory consists of four frames extracted at protocol-level keypoints: \textsc{aspirate}, \textsc{transit}, \textsc{dispense}, and \textsc{post-dispense}. Two NO\_TIP samples are excluded from the quantitative evaluation because YOLO returns no detection on blank tips and the live cascade instead catches those via the separate BLANK\_FLOOR rule (Section~\ref{sec:methods:l2:cascade}), leaving 62 samples for the evaluation reported in Section~\ref{sec:results:l2}.

We report a leakage-free leave-one-plate-out (LOPO) protocol as the canonical evaluation, because it mirrors deployment on a previously unseen plate. For each held-out plate, the PCA world model (Section~\ref{sec:methods:l2:pca}) is fit on the normal trajectories of \emph{all other} plates, the anomaly threshold is calibrated from the held-out plate's own first normal well, and every remaining well of the held-out plate is then scored. No plate ever contributes both to the manifold it is scored against and to its own threshold, so the reported numbers reflect transfer to an unseen plate rather than in-sample reconstruction. 

The primary evaluation uses the p1000 single-channel pipette. To probe the resolution limit of the camera, we additionally collected a matched p20 trajectory set on the same hardware and ran the identical LOPO protocol; the p20 result quantifies where the current camera resolution stops resolving the smaller tip and fluid column (Section~\ref{sec:results:l2}). Extending the pipeline to the 8-channel multichannel head is in progress and reported as future work (Section~\ref{sec:conclusion}).

\subsection{Tip detection (YOLOv8)}
\label{sec:methods:l2:yolo}

\begin{figure}[!t]
  \centering
  \includegraphics[width=0.48\textwidth]{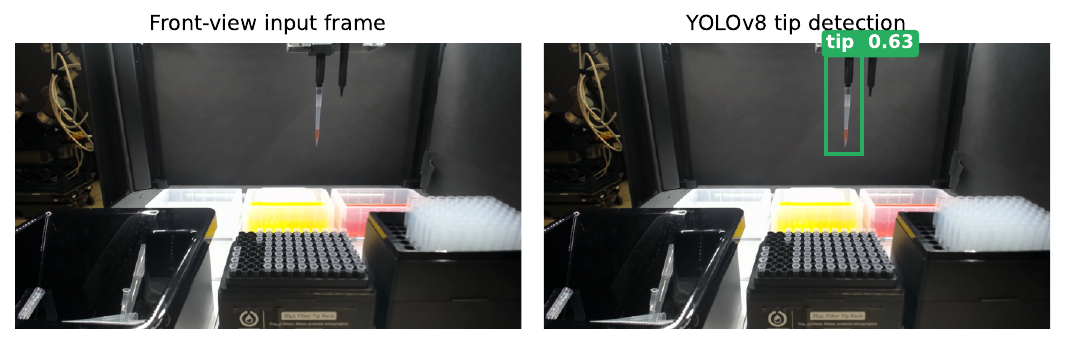}
  \caption{YOLOv8-nano tip detection on a representative camera frame. \textbf{Left:} raw camera input. \textbf{Right:} the fine-tuned detector returns a tight bounding box around the pipette tip (confidence shown above the box). The detected region is cropped, padded, and resized to the fixed canvas used by the PCA world model.}
  \label{fig:yolo-tip}
\end{figure}

To make the per-well pipette region available for downstream modelling, YOLOv8-nano was fine-tuned to localise the pipette tip, on 69 hand-labelled frames in which the tip bounding box was annotated (including tipless frames, where the correct output is no detection). Training took ${\sim}2$ minutes on a single GPU and reached mAP50 of 0.995 on a held-out split. The detector performs geometric tip localisation only and is agnostic to whether a well is normal or failing; interpreting the cropped trajectory is left to the PCA world model and the VLM. At inference, every captured frame yields a tight tip bounding box (Fig.~\ref{fig:yolo-tip}) that is cropped, padded, and resized to a fixed canvas before being passed to the world model.

\subsection{PCA trajectory world model}
\label{sec:methods:l2:pca}

For each well, AEGIS concatenates the YOLO-cropped tip patches from the four keyframes (greyscale, $48\times 48$ each) into one 9{,}216-dimensional trajectory vector. A principal-component analysis with $k{=}15$ components, retaining 97.7\% of variance, is fit on the normal trajectories; $k$ was selected once on the full set of 26 normals, and under the leave-one-plate-out evaluation (Section~\ref{sec:methods:l2:data}) the model is refit per fold. The number of components was chosen at the variance-explained knee, beyond which the reconstruction-error threshold collapses to floating-point noise and the detector degenerates (Section~\ref{sec:results:l2}, Fig.~\ref{fig:pca-components}). Reconstruction error $\|x-\hat{x}\|_2^2$ serves as the anomaly score. Alternative anomaly detectors were evaluated head-to-head against PCA on the same corpus and underperformed at this data scale: an uncalibrated VLM, an Isolation Forest, and a one-class SVM all over-flag the normal wells, and a dense autoencoder learns a near-identity reconstruction at this sample size. The choice of linear PCA is discussed in Results (Section~\ref{sec:results:l2}) and Section~\ref{sec:conclusion}.

\subsection{Real-time deployment: cascade and always-VLM}
\label{sec:methods:l2:cascade}

The reconstruction-error scale drifts by roughly a factor of two between sessions due to lighting, camera profile, and ambient conditions. Rather than re-fit PCA or retune a global threshold, AEGIS calibrates the cascade per run on the first six wells, which by design carry standards, controls, or normalisation reagents that are physically normal. We record their PCA scores $c_1, \ldots, c_6$ and define the silent-pass band as $\mathrm{median}(c_i) \pm 0.7 \cdot \mathrm{MAD}(c_i)$, where MAD is the median absolute deviation. Both the median and the MAD are themselves median operations, so the band is robust to a single anomalous calibration well, and the per-run anchor absorbs cross-session drift without retraining.

\begin{figure}[!t]
  \centering
  \includegraphics[width=0.48\textwidth]{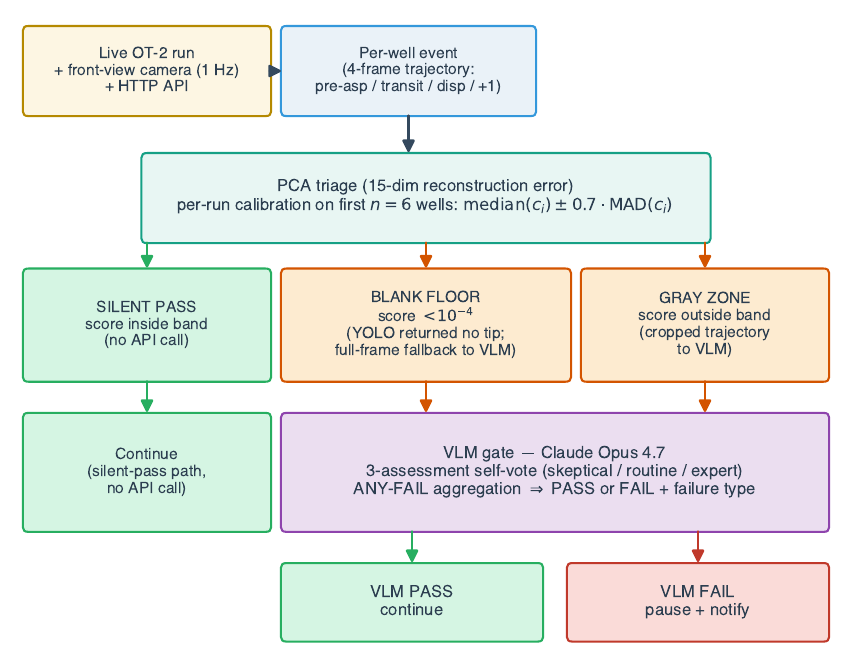}
  \caption{Layer~2 live cascade. Per-well events flow from the camera and the OT-2 HTTP API into PCA triage with three routing zones (SILENT\_PASS inside the per-run calibration band, BLANK\_FLOOR for blank-tip cases, GRAY\_ZONE otherwise). Wells outside the silent-pass band, along with all wells in always-VLM mode, are routed to a 3-assessment self-vote VLM gate that returns the sole pause verdict.}
  \label{fig:cascade-architecture}
\end{figure}

For each post-calibration well, the cascade emits one of three routing decisions (Fig.~\ref{fig:cascade-architecture}):
\begin{itemize}
\setlength\itemsep{0pt}
\item \textsc{Silent-pass} when the PCA score lies inside the per-run calibration band. No API call is issued.
\item \textsc{Blank floor} when the score is below $10^{-4}$, signalling that the YOLO detector produced no tip-containing crop in one or more frames. This is the dominant signature of missing-tip and gross-occlusion failures and is routed to the VLM to confirm.
\item \textsc{Gray zone} for any other score outside the calibration band. The trajectory is routed to the VLM with the YOLO-cropped trajectory frames.
\end{itemize}
The PCA scorer is never permitted to trigger a robot pause on its own. The VLM acts as a second-stage validator that confirms or rejects each PCA-flagged well, and is the sole arbiter of any pause decision: the PCA cascade triages, the VLM decides. This design eliminates a class of false positives in which an unusual but valid well would otherwise halt a working run.

Wells routed by the cascade enter a Claude~Opus~4.7 vision call that receives the four cropped trajectory frames together with a reagent-colour-aware system prompt. The prompt is structured as three in-call assessments by distinct lenses (\textsc{skeptical}, \textsc{routine}, and \textsc{expert}). Each lens independently returns \{\textsc{pass}, \textsc{fail}\} with a confidence and a failure-type label drawn from \{partial dispense, no aspirate, no tip\}. The aggregator applies an \textsc{any-fail} rule: if any lens returns \textsc{fail}, the overall verdict is \textsc{fail}. The reagent-colour parameter enters the prompt explicitly because Claude's prior on coloured fluid presence is strong, and a colour-aware lens is needed to avoid hallucinating ``red fluid'' on an empty tip from a water run.

AEGIS also exposes an always-VLM flag as a second operating point: rather than letting PCA triage, it bypasses the silent-pass band and forces every post-calibration well through the VLM gate. This is a more exhaustive check that always invokes the VLM, trading per-plate cost (Section~\ref{sec:results:demo}) for coverage; it is the safety-maximising setting for cases where one does not want to rely on PCA triage. We stress that always-VLM is not a fix for transparent reagents: water fails in both operating modes (Section~\ref{sec:results:demo}), because the limitation there is one of image resolution rather than of the routing policy, the consumer Logitech webcam places too few pixels on the tip to resolve a transparent meniscus. We therefore treat the two modes only as different points on a cost--coverage curve, and defer the transparent-fluid resolution limit and its potential fixes to Section~\ref{sec:conclusion}.

\label{sec:methods:demo}For the live demonstration the per-robot agent acts on the cascade's pause verdict: when the VLM gate returns \textsc{fail}, the agent issues a pause via the Opentrons HTTP API and the OT-2 halts at the next protocol-level boundary (the drop-tip command at the end of the failed well in our demo protocols). Layered notifications (terminal log, macOS desktop notification, optional Pushover webhook) carry the well identifier, the VLM failure-type label, the lens-by-lens reasoning excerpts, and the path to the saved trajectory frames. Demo protocols exercise the cascade end-to-end on a physical OT-2 with a p1000 single-channel pipette across three physical-failure types (partial dispense, no aspirate, no tip) and three reagent contrasts (red dye, yellow dye, water); a fourth mode, air bubble, could not be tested live because reliably injecting a visible air gap into a running dispense proved difficult, so it is evaluated only in the offline set (Section~\ref{sec:results:l2}). Each injects a single planted failure at a designated transfer and drops the tip after each well so that the cascade's pause arrives before contamination of the next destination. Each condition is run for five consecutive replicates on the same physical OT-2 with a fixed detector configuration, and we report recall as $X/5$ with a Wilson 95\% confidence interval (false pauses are reported separately as raw counts, being uninformative as a proportion at $N{=}5$). No run was excluded on the basis of its outcome.

\section{Results and Discussion}
\label{sec:results}

We report Layer~1 protocol-validation accuracy and ablations (Section~\ref{sec:results:l1}), Layer~2 runtime-monitoring accuracy with baselines and cross-validation (Section~\ref{sec:results:l2}), and the live end-to-end demonstration on a physical OT-2 (Section~\ref{sec:results:demo}). The two layers cover non-overlapping failure surfaces: Layer~1 catches code-level failures invisible to cameras, Layer~2 catches physical-execution failures unpredictable from code, and neither can substitute for the other; AEGIS therefore runs both as separate layers rather than fusing them.

\subsection{Layer 1: protocol validation}
\label{sec:results:l1}

All five LLM backends (o4-mini, the open-weight NVIDIA Nemotron-3 Ultra, Claude Opus~4, GPT-4o, and Claude Sonnet~4.5) classify the assay family correctly on all 24 benchmark protocols (24/24); classification is essentially saturated, with every protocol identified with high confidence from a combination of code structure, variable names, and inline comments.

\begin{figure}[!t]
  \centering
  \includegraphics[width=0.48\textwidth]{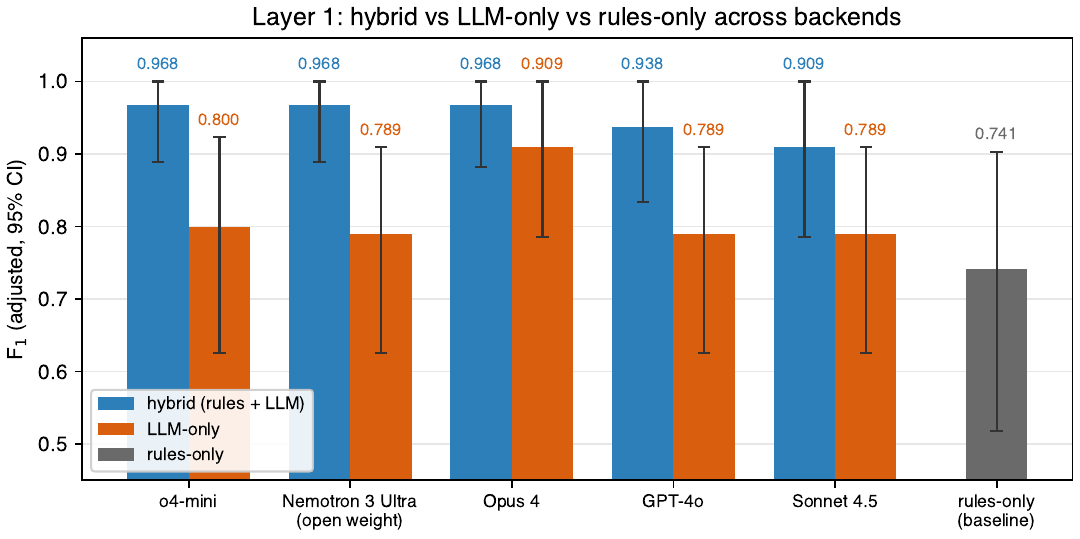}
  \caption{Layer~1 ablation across five backends spanning capability tiers (o4-mini reasoning; frontier Opus~4 and GPT-4o; the mid-tier Sonnet~4.5; and the open-weight Nemotron-3 Ultra) plus the deterministic rules-only baseline, all scored on the same 24 single-bug protocols with 95\% bootstrap confidence intervals on the adjusted F\textsubscript{1}. Each backend shows its hybrid (rules + LLM) and LLM-only bars; rules-only is the grey bar at 0.741. With rules, o4-mini, Opus~4, and the open-weight Nemotron tie at F\textsubscript{1}=0.968 (one false positive each); GPT-4o and Sonnet~4.5 add one and two more. All configurations achieve 100\% recall, so the entire spread is false-positive suppression, and it is supplied by the rules rather than the model.}
  \label{fig:l1-ablation}
\end{figure}

Every LLM-backed configuration also achieves 100\% recall on the 13 injected bugs in the single-bug benchmark: no failure mode in our protocol set evades detection. The five systems differ only in their false-positive rate on the 11 nominally correct protocols. Two of those ``correct'' protocols are Opentrons-tutorial serial-dilution scripts that actually contain a real tip-reuse bug every model legitimately flags; under the adjusted scoring that credits those flags, and with every backend scored on the same 24 single-bug protocols, the five hybrid configurations cluster tightly: o4-mini, the open-weight Nemotron, and Claude Opus~4 all reach F\textsubscript{1}=0.968 (a single false positive each), GPT-4o 0.938 (two false positives), and Claude Sonnet~4.5 0.909 (three), against a deterministic rules-only baseline of 0.741 (Fig.~\ref{fig:l1-ablation}; 95\% bootstrap confidence intervals span roughly $[0.88,1.00]$ for the hybrids). 
The robust observation is that with the rule database present every backend lands within one to three false positives of perfect, regardless of scale, vendor, or cost. The open-model comparison headline follows directly: assay-aware validation does not depend on a paid frontier API, the free open-weight Nemotron ties the best proprietary backend at F\textsubscript{1}=0.968 with 100\% bug recall. The open model's practical price is latency rather than accuracy, roughly $9\times$ slower per protocol than GPT-4o on the free hosted tier ($\sim$57\,s versus $\sim$6\,s), which is acceptable for a pre-flight check that runs once before a multi-hour run.

Stripping the rules exposes each model's intrinsic behaviour, and here the backends diverge sharply. Recall stays at 100\% for all, so once again the difference is entirely false positives, but without the rule database to bound what to check the count jumps from one--three (hybrid) to six--eight. o4-mini falls to F\textsubscript{1}=0.800 (six false positives), and GPT-4o, the open Nemotron, and Sonnet~4.5 to 0.789 (eight); only Claude Opus~4 retains a strong domain prior at 0.909 (three). Without rules GPT-4o flags ten of the eleven correct protocols, essentially refusing to certify anything. Running deterministic regex pattern-matching rules alone (no LLM) misses five bugs, four that require code-flow reasoning (e.g.\ tip reuse through a loop with a default \texttt{new\_tip} parameter, source-reservoir mismatches across function boundaries) and one tutorial tip-reuse variant whose structural pattern the regex does not match, landing the rules-only baseline at 0.741. Every hybrid configuration in Fig.~\ref{fig:l1-ablation} therefore dominates both its own LLM-only ablation and the rules-only baseline.

The false-positive structure carries the main lessons of the Layer~1 comparison. First, the LLM is never the detection bottleneck: every backend catches all injected bugs, so it is the rule database, not the model, that supplies the precision contract. Second, once the rules are present, model scale, vendor, and cost are largely irrelevant, a free open-weight model ties the best proprietary frontier backend, which is the practical result for cost-constrained self-driving laboratories. Third, a larger or newer model is not automatically more precise: Sonnet~4.5 is in fact the weakest hybrid, over-flagging more than Opus, o4-mini, and even the open-weight Nemotron, both with and without rules, so the route to fewer false positives is the rule scaffold rather than a bigger reasoner. Finally, Layer~1's remaining errors are harmless: it occasionally over-flags correct code, but never misses a real bug, which is the safe way for a safety checker to be wrong.

To test whether Layer~1 reports every violation it finds rather than only a pass/fail verdict, we separately evaluated Claude Opus hybrid on three multi-bug protocols carrying 2--3 concurrent injected bugs each. AEGIS detected all 8/8 injected violations and additionally surfaced one legitimate downstream contamination that the injected bugs would cause, with correct per-violation citations in every case.

After the adjustment, Opus's single remaining false positive is on a cherrypicking transfer protocol that uses manual \texttt{pick\_up\_tip()}/\texttt{drop\_tip()} wrapping per transfer, a pattern the LLM conservatively flags as potentially unsafe even though the explicit drops force fresh tips on each operation. We interpret this as the right kind of error to keep: Layer~1 over-cautions on unusual but valid code rather than missing real violations.

The hybrid design is therefore the right architecture for assay-aware protocol validation, and it is graceful in the face of out-of-distribution assays: an LLM run without rules over-flags rather than silently failing, so for any system intended to grow with users' own contributed rules and assay families, where the rule database is incomplete the validator over-cautions rather than under-cautions, and the failure mode of an extension is a known one.

\subsection{Layer 2: PCA trajectory world model}
\label{sec:results:l2}

Layer~2 builds an unsupervised world model of normal pipetting trajectories: a YOLO tip detector localises the pipette in every captured frame, and the four-keyframe tip trajectory (\textsc{aspirate}, \textsc{transit}, \textsc{dispense}, and \textsc{post-dispense}) of each well is projected into a low-dimensional PCA latent fit on normals only. A trajectory that reconstructs poorly against that latent is anomalous by definition; nothing in the training data tells the model what a failure looks like. We report the world-model build (tip detector and PCA), then its evaluation on a held-out trajectory set.

\subsubsection{Tip detection and world-model training}
\label{sec:results:l2:training}

YOLOv8-nano fine-tuned on 69 hand-labelled frames with tip and without reaches mAP50 of 0.995 on a held-out split after $\sim$2 minutes on a single GPU. At inference, every captured frame yields a tight tip bounding box that is cropped, padded, and resized to a fixed 48$\times$48 canvas before downstream scoring.

For each well, the four YOLO-cropped greyscale tip patches (\textsc{aspirate}, \textsc{transit}, \textsc{dispense}, and \textsc{post-dispense}) are concatenated into one 9{,}216-dimensional trajectory vector, and a principal-component analysis is fit on 26 normal p1000 trajectories. The variance-explained scree places the knee at $k{=}15$ components, retaining 97.7\% of normal-trajectory variance; a $k$-sweep over $\{2,5,8,10,12,15,18,20,25\}$ confirms F\textsubscript{1} plateaus at $k{=}15$ and the calibrated reconstruction-error threshold collapses by twelve orders of magnitude between $k{=}15$ and $k{=}25$ as the model interpolates its small training set. Full scree and component-sweep plots are in SI Section~C. Reconstruction error $\|x-\hat{x}\|_2^2$ serves as the per-well anomaly score.

\subsubsection{Evaluation}
\label{sec:results:l2:eval}

\begin{figure}[!t]
  \centering
  \includegraphics[width=0.48\textwidth]{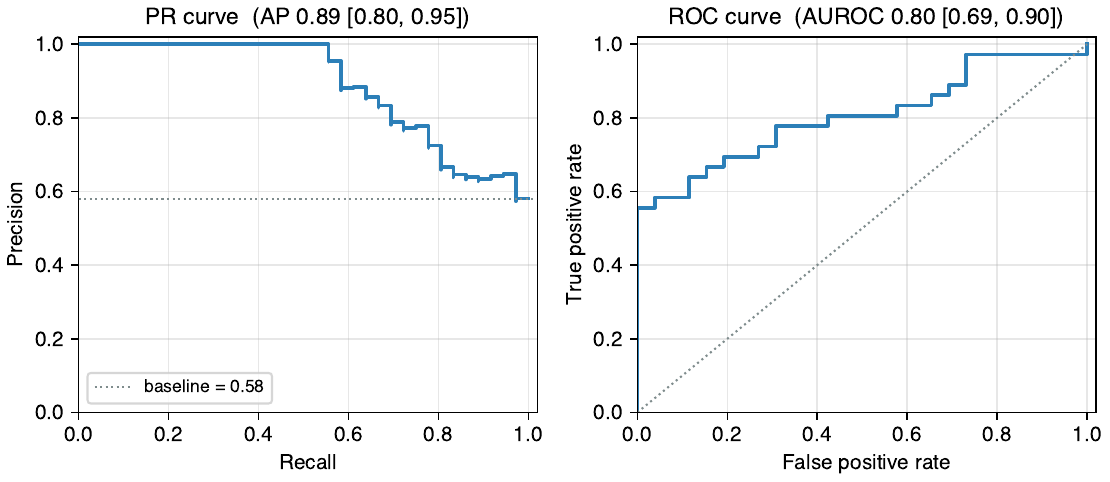}
  \caption{Layer~2 offline detection under the leakage-free leave-one-plate-out protocol on the p1000 set. \textbf{Left:} precision--recall curve (average precision 0.89). \textbf{Right:} ROC curve (AUROC 0.80). The operating point sits at F\textsubscript{1}=0.71 (precision 0.79, recall 0.64) against a failure prevalence of 0.58.}
  \label{fig:l2-lopo}
\end{figure}

We evaluate the world model on 36 planted p1000 failure trajectories, spanning three dye colours (red, yellow, water) and three failure types: partial dispense (single, accumulating, and 25/50/75\% severity variants), no dispense, and air bubble. Under the leave-one-plate-out (LOPO) protocol (Section~\ref{sec:methods:l2:data}), Layer~2 reaches an average precision of 0.887~[0.799, 0.949] and an AUROC of 0.804~[0.685, 0.902] at a failure prevalence of 0.58 (Fig.~\ref{fig:l2-lopo}). At the operating point, F\textsubscript{1}=0.708~[0.567, 0.824], with precision 0.793 and recall 0.639: 23 of the 36 failures caught, at the cost of 6 false positives on the 10 held-out normal wells. All 95\% intervals are well-level bootstrap.

The LOPO result agrees closely with the live chromophoric-cascade recall of ${\sim}0.6$ (Section~\ref{sec:results:demo}), so the offline number predicts what the system achieves in deployment.    

\begin{figure}[!t]
  \centering
  \includegraphics[width=0.48\textwidth]{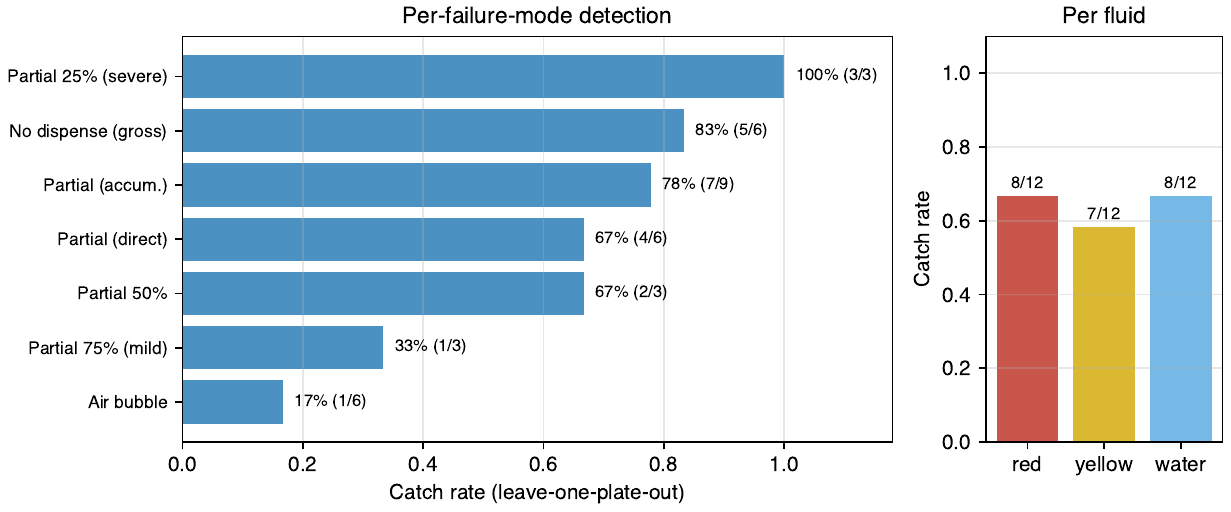}
  \caption{Per-failure-mode and per-fluid catch rate of the trajectory PCA under leave-one-plate-out on the p1000 set. Dispense severity shows a dose-response (25\%/50\%/75\% dispensed $\rightarrow$ 3/3, 2/3, 1/3); air bubble is the weakest mode (1/6). Catch rate is essentially colour-independent (red and water both 8/12), evidence that the scorer keys on tip geometry rather than fluid colour.}
  \label{fig:l2-failure-types}
\end{figure}

The per-failure-mode breakdown (Table~\ref{tbl:l2-perfailure}, Fig.~\ref{fig:l2-failure-types}) shows a clean dose-response: the more liquid a partial dispense leaves in the tip, the more reliably it is caught (25\% dispensed 3/3, 50\% 2/3, 75\% 1/3). No dispense is caught 5/6, and air bubble is the weakest mode at 1/6: a sub-millimetre air gap gives little signal to a side view, and a reliably-visible bubble is itself hard to stage, so some injected gaps may be too small for the camera to resolve. Detection is essentially colour-independent (red and water both 8/12, yellow 7/12) because the scorer keys on the tip meniscus and trajectory geometry, not fluid colour; the geometric signal is present even for water, so the live water miss (Section~\ref{sec:results:demo}) comes from the cascade's tight band and the VLM, not from the detector.

\begin{table}[h]
\small
  \caption{Layer~2 per-failure-type recall under the leakage-free leave-one-plate-out protocol on the p1000 set. The dispense-severity gradient (25/50/75\% dispensed) shows the expected dose-response as the retained-liquid column shrinks; air bubble is the weakest mode.}
  \label{tbl:l2-perfailure}
  \begin{tabular*}{0.48\textwidth}{@{\extracolsep{\fill}}lc}
    \hline
    Failure type & Recall (LOPO) \\
    \hline
    Partial dispense, 25\% dispensed (75\% retained) & 3/3 (100\%) \\
    Partial dispense, 50\% dispensed                 & 2/3 (67\%) \\
    Partial dispense, 75\% dispensed (25\% retained) & 1/3 (33\%) \\
    Partial dispense, single + accumulated           & 11/15 (73\%) \\
    No dispense                                       & 5/6 (83\%) \\
    Air bubble                                        & 1/6 (17\%) \\
    \hline
    \textbf{Total}                                    & \textbf{23/36 (64\%)} \\
    \hline
  \end{tabular*}
\end{table}

\begin{figure}[!t]
  \centering
  \includegraphics[width=0.48\textwidth]{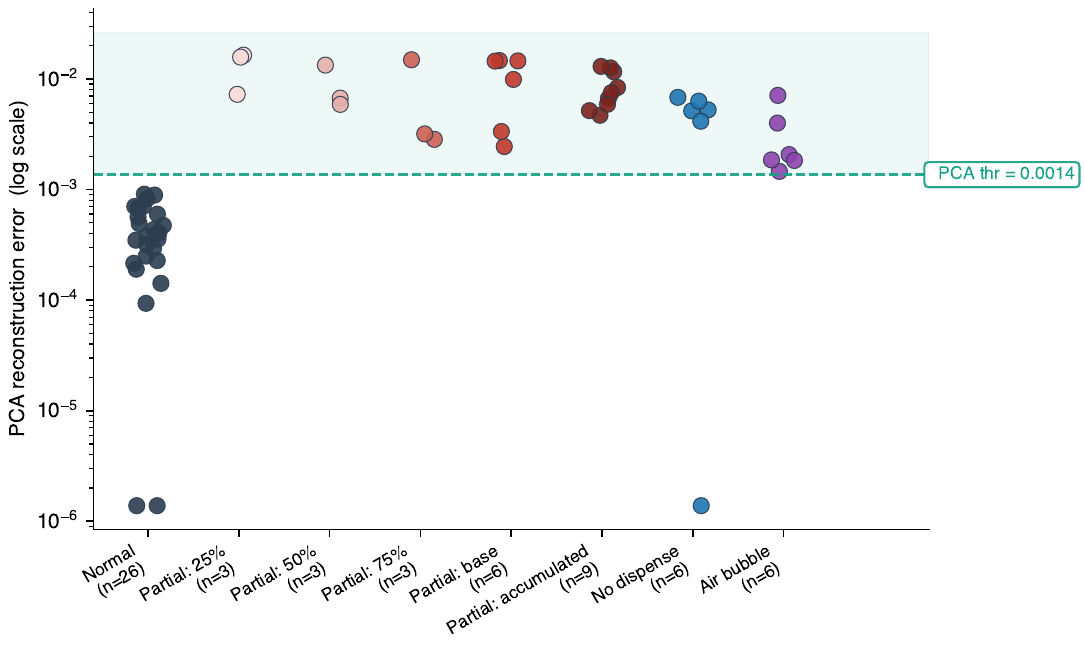}
  \caption{Per-trajectory PCA reconstruction error on the p1000 set ($n{=}62$), grouped by ground-truth class: with no class labels at training time, normal trajectories concentrate at low reconstruction error while the failures (partial-dispense severities, no-dispense, air-bubble) spread toward higher error. The dashed line marks a reference threshold and the shaded band the predicted-failure region; this panel illustrates the signal the detector keys on, with the reported detection numbers coming from the operating point in Fig.~\ref{fig:l2-lopo}. Two NO\_TIP samples are excluded because YOLO returns no detection on blank tips; the live cascade catches those via the BLANK\_FLOOR rule (Section~\ref{sec:methods:l2:cascade}).}
  \label{fig:trajectory-embedding}
\end{figure}

Figure~\ref{fig:trajectory-embedding} shows the raw signal the detector keys on, with no class labels used at training time: normal trajectories sit at low reconstruction error and the failures spread toward higher error. The separation is real but imperfect, and that imperfection is exactly what produces the 6 false positives and 13 misses at the LOPO operating point, because a threshold fixed on one plate's normals does not land perfectly on a new plate.

To choose a scorer, we also compared PCA against an Isolation Forest, a one-class SVM, and an uncalibrated VLM on the same p1000 set. The alternatives over-flag the normal wells whereas PCA does not, and the model-choice argument below explains why a simple linear model is the right fit at this data scale. AEGIS therefore uses linear PCA as its scorer, with a VLM second stage as the cascade's validator (Section~\ref{sec:methods:l2:cascade}).

The choice of a linear model is deliberate. At $n{\sim}26$ normal trajectories in 9{,}216 dimensions, deeper alternatives fail: autoencoders and generative world models (DreamerV3,\cite{DreamerV3} IRIS\cite{IRIS2023}) learn near-identity reconstructions that reproduce failures as faithfully as normals, while density- and boundary-based methods (Isolation Forest, one-class SVM) mis-locate the normal manifold and over-flag. Linear PCA's low-dimensional reconstruction is the right inductive bias for a small, structured training set; train-on-normal PCA with per-run calibration is a strong baseline any guardian system should try before a deeper model.

Per-plate calibration is what makes this transfer across plates. Each plate has its own lighting, camera pose, and reagent colour, and the reconstruction-error scale itself drifts by roughly a factor of two between sessions (the calibration-well mean was $\approx 0.014$ one day and $\approx 0.028$ another). A single fixed threshold therefore transfers poorly and drops precision well below the LOPO value; re-anchoring it to each plate's own first normal well recovers precision to 0.793. That calibration well is free, since any real protocol opens with standards or controls, and it assumes nothing more than that the first six wells of a run are physically normal. The live demonstration was collected in one session and so is free of cross-session drift, but any redeployment on another day or OT-2 will meet it, which is exactly what per-plate calibration absorbs.

The same protocol on the p20 channel marks the resolution limit of a front-view monitor: average precision 0.771, AUROC 0.666, and operating-point F\textsubscript{1} 0.468 (recall 0.306), well below p1000. The p20 tip occupies far fewer pixels, so the meniscus and retained-liquid cues fall near the sensor's floor; the sample-complexity curve is flat in training-set size (SI), confirming a resolution limit rather than a data limit. This bounds where the single-camera setup is trustworthy and motivates a higher-resolution or tip-proximal camera for small-volume channels (Section~\ref{sec:conclusion}).

\subsection{Live demonstration}
\label{sec:results:demo}

\begin{figure}[!t]
  \centering
  \includegraphics[width=0.48\textwidth]{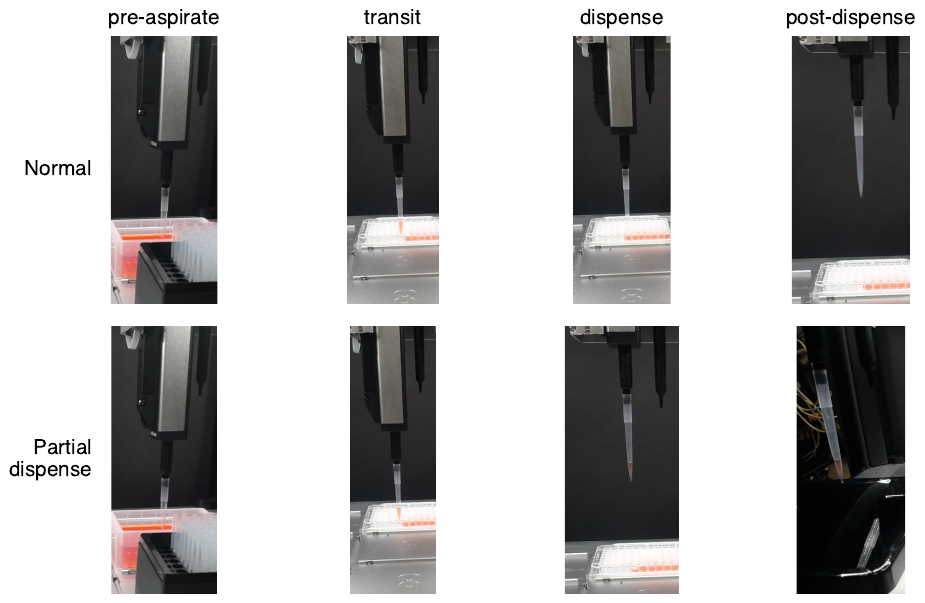}
  \caption{Four-keyframe trajectory representation that feeds the Layer~2 cascade, captured live during a planted partial-dispense demo run on red dye. \textbf{Top row:} a normal well (A7). The pre-aspirate, transit, dispense, and post-dispense crops show a clean empty tip lifting away after a complete dispense. \textbf{Bottom row:} the planted partial-dispense well (A8) from the same run. The post-dispense crop shows the retained dye column still inside the tip after the dispense command completed, which is the dominant visual signal that the PCA scorer and the VLM gate both anchor on.}
  \label{fig:trajectory-keyframes}
\end{figure}

We exercised the cascade end-to-end on a physical OT-2 across the failure modes and three reagent colours of Section~\ref{sec:methods:demo}, five consecutive runs per condition, with recall reported as $X/5$ and a Wilson 95\% interval. We compare two operating points: the default \textsc{cascade} (PCA triage, VLM gating only gray-zone and blank-floor wells) and \textsc{always-VLM} (every post-calibration well gated by the VLM). Table~\ref{tbl:live-demo} gives both, as failure mode $\times$ fluid $\times$ operating point; partial dispense was run on all three fluids, no-aspirate and no-tip on red only. Figure~\ref{fig:trajectory-keyframes} shows the four-keyframe representation for a normal and a planted partial-dispense well, whose retained post-dispense column is the signal that both the PCA scorer and the VLM anchor on.

On chromophoric reagents (red and yellow) the cascade catches partial dispense in 3/5 replicates [0.23, 0.88] and always-VLM lifts this to 5/5 [0.57, 1.00] with unanimous lens agreement, on both dyes. A representative skeptical-lens diagnosis on the red-dye well visualised in Fig.~\ref{fig:trajectory-keyframes}: \emph{``Frame~4 shows a visible red column retained in the tip after dispense and lift-out, indicating incomplete dispense.''}

On water the cascade fails in both directions (0/5 [0.00, 0.43]: one false pause on a normal well, the rest missed), and always-VLM reaches only 1/5 [0.04, 0.62], that catch low-confidence and mislabelled. Every water VLM verdict was medium- or low-confidence, versus high-confidence on the dyes, the signature of an underdetermined side view: a transparent fluid in a transparent tip leaves the VLM almost nothing to read. This is not a system failure but the predicted boundary of any pipette-only front view, and it motivates fusing a top-down plate view for transparent reagents.

On red dye, no-aspirate is a PCA mode, not a VLM mode: the cascade catches 3/5 [0.23, 0.88] from the transit-frame anomaly, but always-VLM \emph{falls} to 2/5 [0.12, 0.77]. An empty tip \emph{after} dispense is ambiguous, a never-filled tip and a normally-emptied one look identical from the side, so the VLM defaults to the benign read, whereas the transit frame is unambiguous. This is the live evidence for keeping PCA triage rather than routing everything to the VLM. No-tip is caught deterministically (5/5 [0.57, 1.00]) by the \textsc{blank floor} rule, since YOLO returns no detection on a tipless shaft; always-VLM is marked n/a because a missing tip already routes to the VLM in either mode, so a separate run would exercise the same path.

\begin{table}[h]
\small
  \caption{Live cascade catches across failure modes, reagent fluids, and operating points (five consecutive replicates per cell; recall $X/5$). \textsc{Cascade} is PCA triage with VLM gating only on gray-zone or blank-floor wells; \textsc{always-VLM} is every post-calibration well routed to the VLM. Partial dispense was exercised on all three fluids; no-aspirate and no-tip were exercised on red dye only. Wilson 95\% intervals: $3/5$ [0.23, 0.88], $5/5$ [0.57, 1.00], $0/5$ [0.00, 0.43], $2/5$ [0.12, 0.77], $1/5$ [0.04, 0.62]. One false pause occurred on a normal well in a water cascade run (Limitations).}
  \label{tbl:live-demo}
  \begin{tabular*}{0.48\textwidth}{@{\extracolsep{\fill}}lcccccc}
    \hline
                     & \multicolumn{3}{c}{Cascade} & \multicolumn{3}{c}{Always-VLM} \\
    \cline{2-4} \cline{5-7}
    Failure mode     & Red & Yellow & Water & Red & Yellow & Water \\
    \hline
    Partial dispense & 3/5 & 3/5    & 0/5   & 5/5 & 5/5    & 1/5   \\
    No aspirate     & 3/5 & --     & --    & 2/5 & --     & --    \\
    No tip           & 5/5 & --     & --    & n/a & --     & --    \\
    \hline
  \end{tabular*}
\end{table}

Always-VLM helps only where the cue is positive (the retained column): it lifts partial dispense to 5/5 but cannot rescue no-aspirate or water. The VLM was the source of the only false pause in the whole demo (a hallucinated failure on a normal water well) and of the no-aspirate misses (anchoring on the legitimately-empty post-dispense frame).

This is why the cascade has its shape: PCA is never allowed to pause on its own (which would over-flag unusual-but-valid wells such as empty controls), and the VLM, unreliable as a standalone scorer, only validates. Neither layer carries a precision contract it cannot meet alone. The 3-assessment self-vote prompt applies the same idea inside one call, an any-fail rule over multiple lenses beat a multi-call temperature ensemble in our tests, because correlated temperature draws suppress minority-correct catches on borderline reagents.

The cascade also exposes a cost dial. A 3-assessment self-vote VLM call costs ${\approx}$\$0.108 per well on Claude Opus~4.7. In the default cascade only ${\sim}15.7\%$ of wells are routed to the VLM, so a 96-well plate costs ${\approx}$\$1.63; always-VLM routes every well at ${\approx}$\$10.33 per plate. The default cascade suits chromophoric reagents where PCA triages reliably; always-VLM is the safety net when it cannot (transparent reagents, low contrast). The break-even scales with reagent contrast, giving the operator an explicit dial rather than one baked-in threshold.

\begin{figure}[!t]
  \centering
  \includegraphics[width=0.48\textwidth]{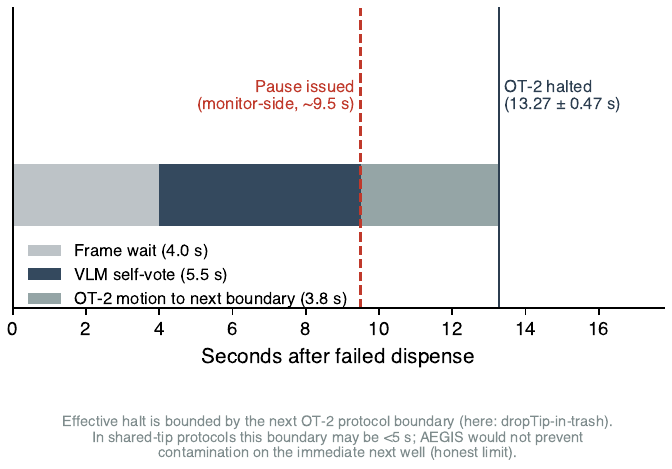}
  \caption{Latency decomposition from failed dispense to OT-2 effective halt. Monitor-side latency (failed dispense to pause API call issued) is $\sim$9.5\,s: a 4\,s frame-flush wait plus a 5.5\,s self-vote VLM call. The OT-2 then applies the pause at the next protocol boundary; effective-halt latency in the demo protocols is $13.27 \pm 0.47$\,s (n=9 paused runs).}
  \label{fig:latency}
\end{figure}

Latency to the OT-2 halt has two parts. The monitor-side latency (failed dispense to pause issued) is deterministic: a 4.0\,s frame-capture wait plus a 5.5\,s VLM call, ${\sim}$9.5\,s. The effective-halt latency (to the robot actually stopping) is set by the next protocol boundary, drop-tip in our demos, and measures $13.27 \pm 0.47$\,s across 9 paused runs (Fig.~\ref{fig:latency}). One honest limit: for protocols that share a tip across consecutive wells, that boundary can arrive before the ${\sim}$9.5\,s monitor latency, so AEGIS would pause but not prevent immediate-next-well contamination; faster triage (signal-driven frame flush, a smaller vision model) is future work.

\subsection{Discussion: findings and limitations}
\label{sec:results:discussion}

\paragraph{Detection is failure-mode-dependent, which is why the cascade has two stages.} No single detector covers every failure. Partial dispense is a VLM strength (the retained column is a direct cue; always-VLM 5/5 on dyes). No-aspirate is the opposite, a PCA mode (cascade 3/5 vs always-VLM 2/5): an empty tip after dispense is ambiguous to the VLM, but the transit frame is not. No-tip is deterministic via BLANK\_FLOOR, which fires only for a genuinely tipless shaft, since YOLO localises an empty-but-present tip (so no-aspirate does not floor). Routing everything to the VLM would lose no-aspirate; trusting PCA alone would forfeit precision, so AEGIS keeps both.

\paragraph{Transparent reagents fail for two reasons, only one a camera problem.} The water result (cascade 0/5, always-VLM 1/5) is a principled limit with two causes that need different fixes. One is geometric: a transparent fluid in a transparent tip, seen from the side, gives almost no contrast, and no sensor fixes that. The other is resolution: the consumer webcam puts too few pixels on the tip, the same bottleneck that caps p20; a higher-resolution or tip-proximal camera helps here and on p20/p300, but not the geometry. The robust fix is a top-down post-run plate view (a prospective Layer~3) that measures dispensed volume independently of tip transparency. The VLM's uniformly low-confidence water verdicts are it reporting this uncertainty honestly rather than guessing.

\paragraph{Reporting choices at small $N$.} We report live recall as $X/5$ with a Wilson interval; a Wald error is inappropriate here, collapsing to zero width at 0/5 and 5/5. The confidence-interval-backed claim remains the offline LOPO number (AP 0.887, F\textsubscript{1} 0.708), with the live matrix the demonstration it predicts. False pauses are reported as raw counts, not a rate, which would be uninformatively wide at $N{=}5$. Per honest-N reporting, the five replicates were consecutive runs with a fixed detector (Section~\ref{sec:methods:demo}) and no run was excluded on its outcome; an unrelated calibration-well false pause was resumed so the planted failure still ran.

\paragraph{The cascade improvement is multi-factor.} We caution against over-attributing the gap between the initial single-replicate cascade and the calibrated design here: the redesign changed three things at once, the calibration count ($n{=}3 \rightarrow 6$), the silent-pass band (raw $[\min, \max] \rightarrow$ robust $\mathrm{median} \pm 0.7\cdot\mathrm{MAD}$), and the fail-well position, so the gain reads as their joint effect, not any one alone. Isolating them is a small future ablation.

\section{Conclusions}
\label{sec:conclusion}

We have presented AEGIS, a two-layer guardian architecture for the Opentrons~OT-2 and similar open-source liquid handlers that addresses the silent-failure problem in self-driving laboratories: failures that occur in either the protocol code or the physical execution and that routinely escape detection until a downstream assay fails or an operator notices an empty well. Layer~1 reads the protocol's Python source before each run and flags assay-level violations against a curated rule database; Layer~2 watches the pipette through an off-the-shelf camera at runtime, scoring per-well trajectories with a small PCA world model and routing uncertain wells to a vision-language model gate that arbitrates pause decisions.

\subsection*{Contributions}

Our primary contribution is the two-layer guardian architecture above, which retrofits onto the existing OT-2 stack with off-the-shelf cameras and no robot-side modifications. The second contribution is the cascade design pattern at Layer~2 itself: PCA-as-triage with per-run calibration, VLM-as-validator, and neither layer permitted to halt a run on its own signal, so that each layer's known failure mode is gated by the other and the operator has an explicit cost-coverage dial between the default and always-VLM operating points. We demonstrate the cascade end-to-end on a physical OT-2 with effective-halt latency $13.27 \pm 0.47$\,s ($n=9$ paused runs), catching planted partial-dispense failures across three reagent contrasts (red, yellow, water) and planted no-aspirate and no-tip failures on red dye, the latter two being independent of reagent identity (the failure signatures are mechanical, not optical). To support reproducibility and community extension, we release under the MIT license a 24-protocol benchmark with planted assay-level bugs (to our knowledge the first public corpus of liquid-handling failures with assay-level rule annotations), a 22-rule assay rule database, the YOLOv8-nano tip-detector weights, the trajectory dataset, the live demo harness, and reproduction scripts for every table and figure in this paper; a separate negative-results record (SI~Section~B) documents alternatives that did not work, recorded so the next group does not retread the same paths.

\subsection*{Key findings and insights}

At Layer~1, all four LLM backends achieve 100\% bug recall on the single-bug benchmark, so the LLM is never the detection bottleneck; the rule database carries the precision contract, and an LLM run without rules over-flags rather than silently failing, which makes the system graceful in the face of out-of-distribution assays. At Layer~2, linear PCA with per-run calibration outperforms both deep generative world models (which overfit on 26 normal trajectories and reconstruct anomalies as faithfully as normals) and boundary or density baselines (Isolation Forest, one-class SVM), a transferable lesson for any guardian trained on a small normal corpus. Per-run $\mathrm{median} \pm 0.7\cdot\mathrm{MAD}$ anchoring further absorbs cross-session drift in the reconstruction-error scale, which we observed to shift by approximately a factor of two between two demonstration sessions on the same hardware; the cascade absorbed this drift without retraining.

Neither layer is trustworthy as a standalone scorer. The PCA can silently pass unusual-but-valid wells; the VLM hallucinates failures on low-contrast crops and misses real failures by anchoring on the wrong frame. The cascade gates each layer's known failure mode through the other and exposes an explicit cost-coverage dial: the default cascade routes only $\sim 15.7\%$ of post-calibration wells to the VLM and costs an estimated \$1.63 per 96-well plate, while the always-VLM operating point routes every well and costs \$10.33 per plate. The two-layer complementarity is architectural by construction: Layer~1 sees the protocol code, Layer~2 sees the physical execution, and neither layer can substitute for the other. Finally, front-view monitoring of transparent reagents is a principled physical limit; AEGIS surfaces this in the VLM's own low-confidence reasoning rather than hallucinating a verdict.

\subsection*{Current limitations}

Five limitations bound the present work. First, the primary evaluation is on the p1000 single-channel pipette; we additionally quantify the p20 channel under the same leave-one-plate-out protocol and find it near the resolution limit of crop-based PCA at our camera setup (F\textsubscript{1} 0.47 versus 0.71 on p1000, with a flat sample-complexity curve indicating a resolution rather than a data limit), while the p300 and the 8-channel multichannel head are likely tractable with the same pipeline but as yet untested (8-channel runs are in progress). Second, the n=6 per-run calibration with a $\mathrm{median} \pm 0.7\cdot\mathrm{MAD}$ band is robust to a single anomalous calibration well but not to a cluster of mis-behaving calibration wells; cross-run normal priors that accumulate as more sessions log their data would shrink this variance further. Third, front-view monitoring of transparent reagents is a principled limit on volume-based failures: a transparent fluid in a transparent tip is nearly invisible from the side. A higher-resolution or tip-proximal camera would help the resolution component of this limit but not the underlying geometry, so a top-down post-run plate view (Layer~3) is the robust fix. Fourth, the monitor-side latency of $\sim$9.5\,s can exceed the next protocol-level boundary in protocols that share a tip across consecutive destination wells; AEGIS still halts the run but would not prevent immediate-next-well contamination in those cases. Fifth, the live demonstration is on a single OT-2 across three planted-failure modes and three reagent contrasts; air-bubble injection proved difficult to stage reliably in a live run and so could not be tested live (it is evaluated only in the offline set, where it is the weakest mode), and coverage of new assays and multi-protocol scheduling on a heterogeneous fleet remain future work.

\subsection*{Future work}

The most important future direction is strengthening the live cascade so that failures are caught the moment they occur, reducing both the post-run audit burden and the wasted reagent and time that result when a failure is only detected after a long run completes. As more normal trajectories accumulate across deployments, a stronger world model (a fine-tuned successor to linear PCA, or an online streaming PCA that adapts continuously and removes the explicit calibration phase) will absorb more of the routing decision, the gray-zone fraction will shrink, and the per-plate VLM call count will drop without changing the cascade's safety contract. A tiered VLM gate where a smaller and faster vision model triages the gray-zone and the heavier model arbitrates only the ambiguous remainder is feasible today and cuts the current $\sim$9.5\,s monitor-side floor by an estimated 3--5$\times$ on the routine majority of paused wells. On the sensing side, an active camera that zooms or pans to track the tip in real time, steered by the existing YOLO tip detector, would place far more pixels on the tip and directly attack the resolution limit behind both the p20 small-pipette and transparent-reagent cases.

More broadly, the linear PCA scorer is a deliberate baseline, not a ceiling: it is the right model at $n{\sim}26$ normal trajectories, where deeper alternatives overfit, but it is the component most likely to improve as data grows. With more normal trajectories and, crucially, more labelled failures, a fine-tuned autoencoder or deep world model, or a foundation or vision-language model fine-tuned on this corpus, should raise detection on exactly the modes PCA handles least well today (air bubble, transparent reagents, small pipettes), while the train-on-normal, per-plate-calibrated cascade around it stays the same. AEGIS is thus a data-efficient starting point whose accuracy scales with the datasets its own deployment generates.

In the meantime, a post-run audit layer (L3) is a near-term complement that re-scores every well after the run with a slower-and-stricter ensemble and surfaces silent-passes for human review. A prototype already caught the water partial-dispense case retrospectively, and the role of the audit naturally diminishes as the live cascade strengthens.

The longest-term extension is closing the loop from detection to autonomous correction, with two distinct cases. Some failures are recoverable at the protocol level: re-aspirating from a fresh reservoir, swapping tips from a remaining tip rack, or retrying a transfer. For these, an AEGIS agent with sufficient context could trigger the correction itself or hand control to a protocol-rewriting predictive model that proposes a corrective sub-protocol on the fly. Hardware-level failures (a missing tip on the rack, a clogged tip mid-run, deck contamination, a knocked-over reservoir) are outside the OT-2's own actuation envelope, so true autonomy in those cases would require a physical-intervention layer such as a humanoid arm or robotic operator that loads tips, replaces consumables, and clears mechanical faults. AEGIS as the perception layer fits naturally into either trajectory: the same trajectory features and VLM verdicts that pause the run today could route to a corrective sub-protocol for the software-recoverable class, or to a physical-intervention request for the hardware-level class.

Two narrower directions follow from the Layer~1 and benchmark sides. The Layer~1 validator is currently a single LLM call per protocol; a multi-call ensemble is the obvious lever for further reducing false positives at higher cost, though we did not pursue it here because the Layer~2 VLM-gate analogue (multi-call temperature ensemble) produced correlated errors that majority-vote suppressed on minority-correct catches in our small evaluation. The released benchmark invites community contributions in both data and rules; fine-tuning smaller open-source models on this corpus would lower per-protocol API cost and bring both layers into reach for fully-local deployment.

\subsection*{Broader impact}

The two failure surfaces AEGIS targets occur in every open-source self-driving laboratory we are aware of, and they routinely escape detection until a downstream assay fails or an operator notices an empty well. Defense in depth across non-overlapping surfaces, namely code-level validation and physical-execution monitoring, reduces the silent-failure rate that otherwise compounds across runs in an autonomous lab. The cascade pattern itself, cheap triage with an expensive validator and neither layer acting alone, generalises beyond AEGIS to any deployment where a strong but imperfect cheap model and a more capable but fallible LLM gate are both available. By releasing the benchmark, the rules, the YOLO weights, the trajectory dataset, and the live demo harness under a permissive license, we lower the floor for adding a guardian layer to any community OT-2 setup; there is little reason for a new open-source self-driving laboratory to operate fully open-loop.

\section*{Author contributions}

\section*{Conflicts of interest}
There are no conflicts to declare.

\section*{Data availability}
The AEGIS source code, protocol benchmark (24 OT-2 protocols across 5 assay families), assay rule database, Layer~2 trajectory dataset (128 labelled samples across 3 reagent contrasts and 3 physical failure types), trained YOLOv8 weights, and reproduction scripts are available at \url{[GitHub URL to be added on release]} under the MIT license. Raw camera captures are deposited at [Zenodo DOI to be assigned upon acceptance].

\section*{Acknowledgements}
We thank the Opentrons community for protocol examples used in our correct-protocol benchmark, and the Academy framework team at Globus Labs for early access and integration support. This work was supported by [funding sources to be added]. Argonne National Laboratory's work was supported by the U.S. Department of Energy, Office of Science, under contract DE-AC02-06CH11357.



\balance

\bibliography{references}
\bibliographystyle{rsc}


\appendix

\section{Supplementary Information}
\label{sec:appendix}

\subsection{A. Multi-robot fleet deployment via Academy}
\label{sec:si:fleet}

\paragraph{Architecture.~~} To support multi-robot deployment, AEGIS wraps both layers as a single \texttt{AEGISAgent} class (\texttt{aegis/agents/aegis\_agent.py}) exposing one \texttt{@action process\_protocol(\dots)} entry point. The agent owns one \texttt{ProtocolValidator} (Layer~1) and one \texttt{Layer2Inference} (Layer~2) per robot. A fleet of $N$ such agents is launched under \texttt{academy.manager.Manager}\cite{Academy2025} with the \texttt{LocalExchangeFactory} and a \texttt{ThreadPoolExecutor}; protocols are partitioned across agents and processed in parallel. No code in either layer is aware of the fleet: scaling AEGIS is launching more agents.

\begin{figure}[h]
  \centering
  \includegraphics[width=0.48\textwidth]{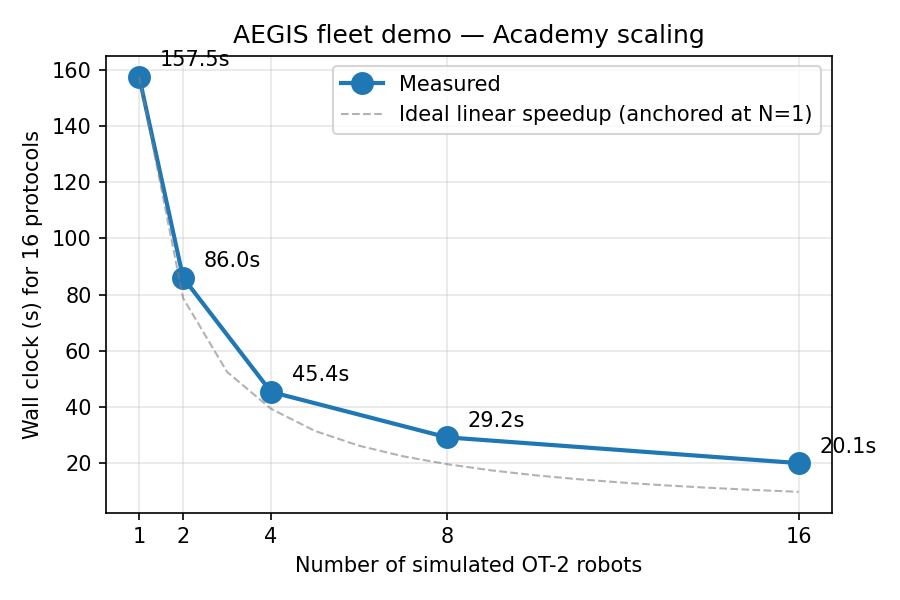}
  \caption{Academy fleet scaling on the 16-protocol benchmark using simulated OT-2 backends. Wall-clock time drops $7.84\times$ from $N{=}1$ to $N{=}16$ agents; detection is invariant at 15/16 across the sweep. Sublinearity at high $N$ is floored by per-protocol Layer~1 LLM latency.}
  \label{fig:academy-scaling}
\end{figure}

\paragraph{Scaling sweep (simulated fleet).~~} We swept $N \in \{1, 2, 4, 8, 16\}$ \texttt{AEGISAgent} instances on a 16-protocol benchmark workload using simulated OT-2 backends. Detection is invariant at 15/16 across the sweep; wall-clock time drops $7.84\times$ from $N{=}1$ to $N{=}16$ (Table~\ref{tbl:fleet}, Fig.~\ref{fig:academy-scaling}). Sublinearity at high $N$ is gated by per-protocol Layer~1 latency (single LLM call, ${\sim}8$--$10$\,s on Opus): once each agent has only one protocol to process, the slowest single LLM call floors the wall clock. The architectural implication is that scaling AEGIS to a real heterogeneous fleet is a deployment exercise rather than a redesign; the per-robot agent already encapsulates the necessary state. The scaling sweep is on simulated robots and does not characterise heterogeneous-hardware behaviour, which we list as a future-work item in the main text.

\begin{table}[h]
\small
  \caption{Academy fleet-scaling sweep on the 16-protocol benchmark, simulated OT-2 backends. Detection is L1$+$L2 verdict agreement on each protocol.}
  \label{tbl:fleet}
  \begin{tabular*}{0.48\textwidth}{@{\extracolsep{\fill}}ccccc}
    \hline
    $N$ robots & Wall (s) & Speedup & \% ideal & Detection \\
    \hline
    1  & 157.5 & 1.00$\times$ & 100\% & 15/16 \\
    2  & 86.0  & 1.83$\times$ & 92\%  & 15/16 \\
    4  & 45.4  & 3.47$\times$ & 87\%  & 15/16 \\
    8  & 29.2  & 5.39$\times$ & 67\%  & 15/16 \\
    16 & 20.1  & \textbf{7.84$\times$} & 49\%  & 15/16 \\
    \hline
  \end{tabular*}
\end{table}

\subsection{B. Negative results: alternatives that did not work}
\label{sec:si:negatives}

In the course of building Layer~2 we evaluated several alternatives that did not work and are documented here for completeness:

\begin{itemize}
\setlength\itemsep{0pt}
\item \emph{Multi-call VLM ensemble with temperature sampling.} Three independent VLM calls with $T{>}0$ and majority vote over their verdicts. The three draws produced highly correlated errors on borderline cases; majority vote then suppressed minority-correct catches. The in-prompt 3-assessment self-vote that we use in the live cascade resolves this by aggregating with an \textsc{any-fail} rule across distinct lenses inside a single call.
\item \emph{Post-dispense single-frame PCA.} Scoring only the post-dispense frame (rather than the four-frame trajectory) cannot detect failures in which the tip is empty in both normal and failure conditions (no aspirate, no dispense, air bubble during transit). The four-frame trajectory representation supplies the temporal context that the single-frame view lacks.
\item \emph{Convolutional autoencoder world model.} Even with aggressive bottlenecks, autoencoders trained on 26 normal trajectories learn near-identity reconstructions and reconstruct failure trajectories as faithfully as normals, vanishing the discriminator. Linear PCA's lower-capacity inductive bias is the right match for the small-sample regime.
\item \emph{Crop-based PCA on the p20 pipette.} The 20\,\textmu L pipette tip is geometrically thin at our camera setup, so cropped trajectory vectors of failure and normal wells are only weakly separable: the same leave-one-plate-out protocol yields average precision 0.77, AUROC 0.67, and operating-point F\textsubscript{1} 0.47 (recall 0.31), well below the p1000 numbers (0.89 / 0.80 / 0.71). The sample-complexity curve is flat in the number of training normals, indicating a resolution rather than a data limit. A higher-resolution front view or a different representation (e.g., a VLM-only path, or a specialised meniscus detector) is the natural next step.
\item \emph{Top-down iPhone capture for in-flight monitoring.} The OT-2 gantry occludes the top-down plate view at most pipetting moments; even after homing, the gantry's resting position can block portions of the plate. We retain the top-down capture only for post-run plate state and document the in-flight occlusion as a deck-geometry limitation.
\item \emph{PCA-projected Mahalanobis (15-dim).} Computing Mahalanobis distance in the 15-dimensional PCA-projected space (rather than the 32-dimensional hand-crafted feature space) fails because failures project to lower-energy regions of the PCA latent than normals; the very directions PCA selects are the directions in which failures concentrate close to the origin, inverting the distance metric.
\end{itemize}

\subsection{C. PCA component selection for the trajectory world model}
\label{sec:si:pca-components}

\begin{figure}[H]
  \centering
  \includegraphics[width=0.48\textwidth]{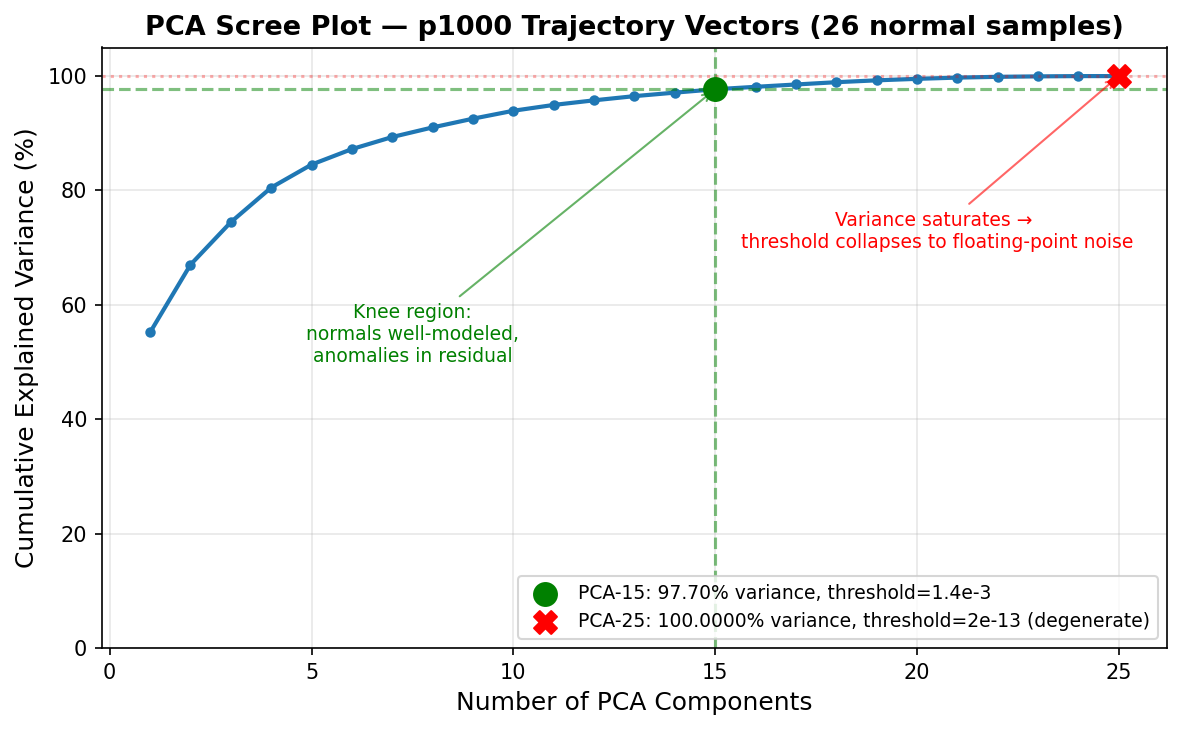}
  \caption{Variance-explained scree for the trajectory PCA. The knee at $k{=}15$ retains 97.7\% of normal variance; beyond $k{=}20$ the model interpolates its 26-trajectory training set and the calibrated reconstruction-error threshold collapses to floating-point noise.}
  \label{fig:pca-scree}
\end{figure}

\begin{figure}[H]
  \centering
  \includegraphics[width=0.48\textwidth]{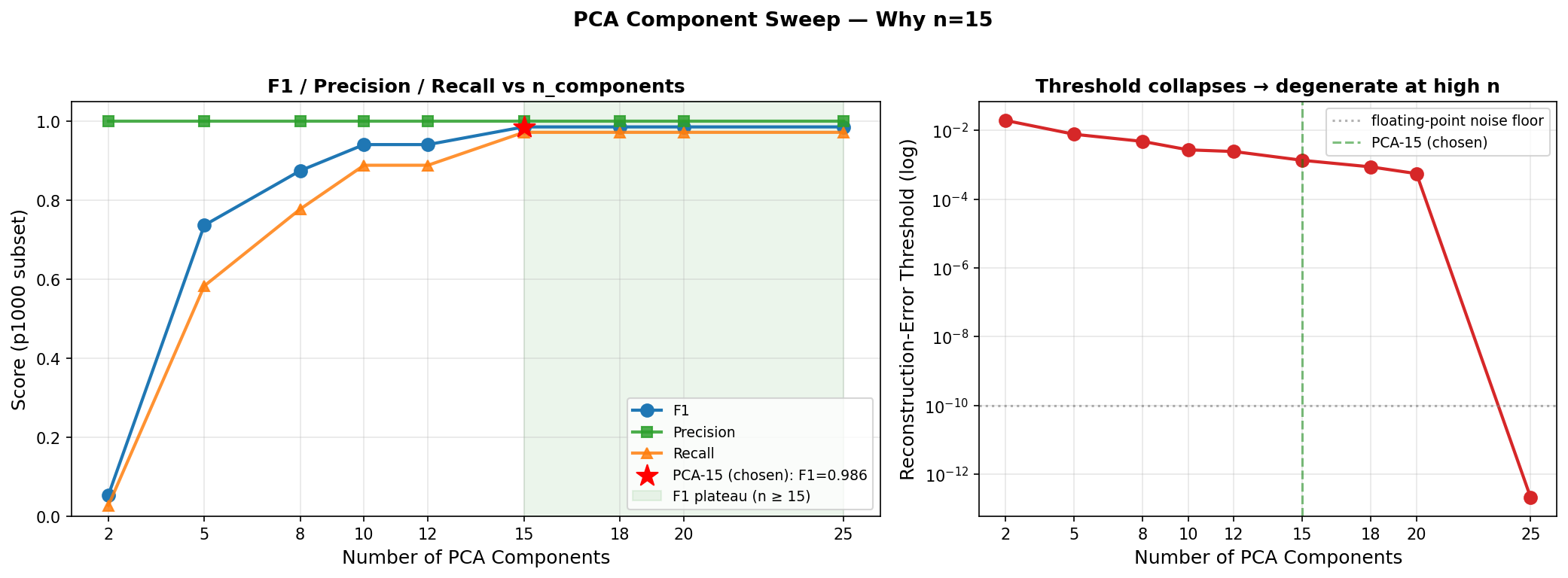}
  \caption{PCA component sweep. F\textsubscript{1}, precision, and recall on the p1000 set across $k$, alongside the calibrated threshold. F\textsubscript{1} plateaus at $k{=}15$; the threshold collapses at high $k$ as the model interpolates the training set.}
  \label{fig:pca-components}
\end{figure}

The PCA component count $k$ for the trajectory world model is chosen at the variance-explained knee, where F\textsubscript{1} on the held-out set plateaus and the calibrated reconstruction-error threshold has not yet collapsed into floating-point noise. The variance-explained scree (Fig.~\ref{fig:pca-scree}) places the knee at $k{=}15$, retaining 97.7\% of normal-trajectory variance. A $k$-sweep over $\{2,5,8,10,12,15,18,20,25\}$ confirms F\textsubscript{1} rises monotonically up to $k{=}15$ and plateaus thereafter, while the calibrated threshold collapses by roughly twelve orders of magnitude between $k{=}15$ and $k{=}25$ as the model begins to interpolate its 26-trajectory training set (Fig.~\ref{fig:pca-components}). Beyond the knee the detector degenerates: reconstruction error of both normals and failures shrinks to floating-point noise and the discriminator vanishes.

\end{document}